\documentclass{article}
\usepackage{spconf,amsmath,graphicx}

\usepackage[english]{babel}
\usepackage[T1]{fontenc}  % choose the output font encoding 
\usepackage[latin1]{inputenc} % allows the user to input accented characters
\usepackage{url}
\usepackage{graphicx}
\usepackage{subfigure}
\usepackage{xcolor}
\usepackage{amsmath,amsfonts,amssymb}
\usepackage{booktabs}
\usepackage{epsfig}
\usepackage{pstricks}
\usepackage{pst-node}
\usepackage{pst-grad}
\usepackage{ifthen}
\usepackage{algorithm}
\usepackage{algorithmic}	% spf add 18/09/2015
\usepackage{setspace}           % For commands \singlespacing, \onehalfspacing, \doublespacing

\usepackage{epstopdf}	% spf add 22/05/2015
\usepackage{enumerate} % spf add 28/07/2015
\usepackage{multicol} % split the algorithm into two columns
\usepackage{enumitem} % customize a list environment, such as itemize, enumerate
\usepackage{cite} % add hyphen between multiple reference
\usepackage{appendix}

\allowdisplaybreaks

\newcommand{\mypar}[1]{\bigskip\noindent {\bf #1.}}

\definecolor{red}{RGB}{153,0,0}

\title{Multi-modal Image Processing based on Coupled Dictionary Learning}

\name{
	Pingfan Song$^{\star}$ \quad
	Miguel R.\ D.\ Rodrigues$^{\star}$  
	\thanks{
		This work was supported by the Royal Society International Exchange Scheme IE160348, by UCL Overseas Research Scholarship (UCL-ORS) and by China Scholarship Council (CSC).
	}
}
\address{				
	$^{\star}$ Department of Electronic and Electrical Engineering, University College London, UK%
}

\begin{document}
%\ninept % use 9 pt font, otherwise use 10 pt font
%% make the title area
\maketitle

\begin{abstract}

	In real-world scenarios, many data processing problems often involve heterogeneous images associated with different imaging modalities. Since these multimodal images originate from the same phenomenon, it is realistic to assume that they share common attributes or characteristics.
	In this paper, we propose a multi-modal image processing framework based on coupled dictionary learning to capture similarities and disparities between different image modalities. In particular, our framework can capture favorable structure similarities across different image modalities such as edges, corners, and other elementary primitives in a learned sparse transform domain, instead of the original pixel domain, that can be used to improve a number of image processing tasks such as denoising, inpainting, or super-resolution. Practical experiments demonstrate that incorporating multimodal information using our framework brings notable benefits.	

\end{abstract}

\section{Introduction}
\label{sec:Introduction}

\vspace{-0.2cm}

%Typical image processing tasks focus mainly on single modality images without exploiting the availability of other modalities for aid. However, 

%
In many practical application scenarios, it is common to image a certain scene using various sensors that yield different image modalities. For example, in remote sensing domain, it is typical to have various image modalities of earth observations, such as a panchromatic band version, a multispectral bands version, and an infrared (IR) band version~\cite{gomez2015multimodal,loncan2015hyperspectral}. These different bands often exhibit similar textures, edges, corners, boundaries, or other salient features. In medical imaging domain, multi-contrast scans for the same underlying anatomy~\cite{cherry2006multimodality,townsend2008multimodality,catana2013pet}, such as simultaneous positron emission tomography (PET) / magnetic resonance imaging (MRI) scans, MRI T1/T2-weighted scans, also indicate strong correlation. In colorization\cite{levin2004colorization} tasks, the output image has both chrominance channels and luminance channel which share consistent edges. These scenarios call for approaches that can capitalize on the availability of multiple image modalities of the same scene, object, or phenomenon to address interested image processing tasks.

A number of multimodal image processing approaches have also been proposed in the literature to capitalize on the availability of additional \emph{guidance} or \emph{side information}~\cite{renna2016classification,mota2017compressed}, to aid the processing of target modalities, also referred as joint/collaborative image filtering~\cite{kopf2007joint,he2013guided,ham2017robust,shen2015multispectral,li2016deep,zhang2014rolling,wang2012semi,zhuang2013supervised,liu2014semi,jing2015super,dao2016collaborative,bahrampour2016multimodal,deligiannis2017multi,deligiannis2016Xray}. Generally, the basic idea behind these approaches is that the structural details of the guidance image can be transferred to the target image. These approaches have been applied to multi-modal image denoising, super-resolution, classification, and more. 
%For example, it is favorable to purify the noisy image modality with the aid of the clean images of a different modality, or enhance the blurred low-resolution image via exploiting another high-quality modality. 
However, these methods tend to introduce notable texture-copying artifacts, i.e. erroneous structure details that are not originally present in the target image because such methods typically fail to distinguish similarities from disparities between different image modalities. 

In this paper, we propose a new multimodal image processing framework based on coupled dictionary learning. In particular, our model captures complex correlation between different modalities using joint sparse representations over a set of adaptive coupled dictionaries. This enables us to take into account both similarities and disparities of different modalities via considering their common and unique sparse representations which are obtained in learned sparse domains. This characteristic makes our approach robust to inconsistencies between the guidance and target images, as proper guidance information can be extracted from the guidance modality and then be incorporated in a task-specific formulation to aid the processing of the target modality. Moreover, due to the sparsity prior, our model also demonstrates better robustness than other competing methods in presence of noise.

\vspace{-0.3cm}

\section{Multi-modal Image Denoising}
\label{sec:SIMIS}

\vspace{-0.2cm}

We first present our coupled dictionary learning framework for multi-modal image denoising.

\vspace{-0.2cm}

\subsection{Data Model for Denoising}

\vspace{-0.2cm}

%By capitalizing on this model, we propose a novel coupled image denoising scheme. Assuming that the coupled dictionaries associated with the model in \eqref{Eq:SparseModelX}, \eqref{Eq:SparseModelY} and \eqref{Eq:SparseModelX_low} have been learned, and we are now 

Consider a vectorized noisy image $\mathbf{X}^{ns} \in \mathbb{R}^{N}$ of one modality and a corresponding registered clean vectorized guidance image $\mathbf{Y} \in \mathbb{R}^{N}$ of different modality as side information. We first extract (overlapping) image patch pairs from this pair of multimodal images. In particular, let $\mathbf{x}^{ns}_{i} = \mathbf{R}_i \mathbf{X}^{ns} \in \mathbb{R}^{n}$ denote the $i$-th noisy image patch extracted from $\mathbf{X}^{ns}$ and let $\mathbf{y}_{i} = \mathbf{R}_i \mathbf{Y} \in \mathbb{R}^{n}$ denote the corresponding $i$-th clean guidance image patch extracted from $\mathbf{Y}$, where the matrix $\mathbf{R}_i$ is an $n \times N$ binary matrix that extracts the $i$-th patch from the image. Then, we propose a data model to capture the relationship -- including similarities and disparities -- between the two different modalities as follows:
%Accordingly, the matrix $\mathbf{R}_i^T$ is its transposed version that places the $i$-th patch at its proper position in the image.
\begin{align}
\mathbf{x}^{ns}_i 
&= 
\boldsymbol{\Psi}_{c} \, \mathbf{z}_i + \boldsymbol{\Psi} \, \mathbf{u}_i + \epsilon \,,
\label{Eq:SparseModelX_Noise}
\\
\mathbf{y}_i
&= 
\boldsymbol{\Phi}_{c} \, \mathbf{z}_i + \boldsymbol{\Phi} \, \mathbf{v}_i \,,
%\gamma \cdot \mathbf{y}_i
%&= \gamma \cdot \mathbf{R}_i \mathbf{Y}
%= \boldsymbol{\Phi}_{c} \, \mathbf{z}_i + \boldsymbol{\Phi} \, \mathbf{v}_i \,,
\label{Eq:SparseModelY_Noise}
\end{align}
where the sparse representation $\mathbf{z}_i \in \mathbb{R}^{K}$ is common to both modalities, the sparse representation $\mathbf{u}_i \in \mathbb{R}^{K}$ is specific to modality $\mathbf{x}$, while the sparse representation $\mathbf{v}_i \in \mathbb{R}^{K}$ is specific to modality $\mathbf{y}$. In turn, $\boldsymbol{\Psi}_{c} $ and $\boldsymbol{\Phi}_{c} \in \mathbb{R}^{n \times K}$ are a pair of dictionaries associated with the common sparse representation $\mathbf{z}_i$, whereas $\boldsymbol{\Psi}$ and $\boldsymbol{\Phi} \in \mathbb{R}^{n \times K}$ are dictionaries associated with the specific sparse representations $\mathbf{u}_i$ and $\mathbf{v}_i$, respectively. 
Note that the common sparse representation $\mathbf{z}_i$ connects the patches of the two different modalities. The disparities between modalities $\mathbf{x}$ and $\mathbf{y}$ are distinguished by the sparse representations $\mathbf{u}_i$ and $\mathbf{v}_i$, respectively. 
Finally, $\epsilon \in \mathbb{R}^n$ denotes additive zero-mean and homogeneous white Gaussian noise with the standard deviation $\sigma$.

%$\boldsymbol{\Psi}_{c} = [\psi_{c1}, \cdots, \psi_{ck}] \in \mathbb{R}^{n \times k}$ and $\boldsymbol{\Phi}_{c} = [\phi_{c1}, \cdots, \phi_{ck}] \in \mathbb{R}^{n \times k}$ $\boldsymbol{\Psi} = [\psi_{1}, \cdots, \psi_{K}] \in \mathbb{R}^{n \times k}$ and $\boldsymbol{\Phi} = [\phi_{1}, \cdots, \phi_{K}] \in \mathbb{R}^{n \times k}$

\vspace{-0.2cm}

\subsection{Coupled Image Denoising}
%Assume that the coupled dictionaries $\boldsymbol{\Psi}_{c} $, $\boldsymbol{\Phi}_{c}, \boldsymbol{\Psi}, \boldsymbol{\Phi}$ have been learned

Our coupled image denoising problem is addressed in two steps: coupled dictionary learning and reconstruction of the denoised image.

\mypar{Step 1: Coupled dictionary learning}
In this stage, given a set of training patch pairs $\{\mathbf{x}^{ns}_{i}, \mathbf{y}_{i} \}$, we aim to solve the following optimization problem:
%%
%\begin{equation} \label{Eq:Lasso}
%\begin{array}{cl}
%\underset{\mathbf{z}_i,\mathbf{u}_i,\mathbf{v}_i}{\text{min}} 
%&  \! \! \! \! \!
%\left\|
%\begin{bmatrix} 
%\mathbf{x}^{ns}_{i} \\ \mathbf{y}_{i}
%%\mathbf{x}^{ns}_{i} \\ \gamma \, \mathbf{y}_{i}
%\end{bmatrix}
%-
%\begin{bmatrix}
%\boldsymbol{\Psi}_{c} & \boldsymbol{\Psi} & \mathbf{0} \\
%\boldsymbol{\Phi}_{c} & \mathbf{0} & \boldsymbol{\Phi} \\
%\end{bmatrix}
%%
%\begin{bmatrix}
%\mathbf{z}_i \\
%\mathbf{u}_i \\
%\mathbf{v}_i \\
%\end{bmatrix}
%\right\|_2^2
%+
%\lambda
%\left\| 
%\begin{bmatrix}
%\mathbf{z}_i \\
%\mathbf{u}_i \\
%\mathbf{v}_i \\
%\end{bmatrix} 
%\right\|_0
%\end{array}
%\end{equation}
%
%\noindent
%According to the theory of the Lagrangian multipliers, Problem \eqref{Eq:Lasso} can also be written in an equivalent constrained form
%% sparsity constraint version for joint denoising
\begin{equation} \label{Eq:CDLDN}
\footnotesize %\footnotesize %\scriptsize
%\begin{array}{cl}
\underset{\mathbf{D}, \boldsymbol{\alpha}
%	\begin{subarray}{c}
%	\boldsymbol{\Psi}_{c}, \boldsymbol{\Psi}, \boldsymbol{\Phi}_{c}  \\
%	\boldsymbol{\Phi}, \mathbf{z}_i, \mathbf{u}_i,\mathbf{v}_i
%	\end{subarray}
}{\min} 
\,
\sum\limits_i
\left\|
\begin{bmatrix} 
\mathbf{x}^{ns}_{i} \\ \mathbf{y}_{i}
%\mathbf{x}^{ns}_{i} \\ \gamma \, \mathbf{y}_{i}
\end{bmatrix}
-
\begin{bmatrix}
\boldsymbol{\Psi}_{c} \; \boldsymbol{\Psi} \;\, \mathbf{0} \\
\boldsymbol{\Phi}_{c} \;\; \mathbf{0} \;\; \boldsymbol{\Phi} \\
\end{bmatrix}
\begin{bmatrix}
\mathbf{z}_i \\
\mathbf{u}_i \\
\mathbf{v}_i \\
\end{bmatrix}
\right\|_2^2
\,
%\\
\textrm{s.t. }
\, \! \!
%&  \! \! \! \! \!
\left\| 
\begin{bmatrix}
\mathbf{z}_i \\
\mathbf{u}_i \\
\mathbf{v}_i \\
\end{bmatrix} 
\right\|_0
\leq s_i
\;, \forall i,
%\end{array}
\end{equation}
%%% error constraint version for joint denoising
%\begin{equation} \label{Eq:CDLDN}
%\footnotesize %\footnotesize %\scriptsize
%%\begin{array}{cl}
%\underset{
%	\begin{subarray}{c}
%	\boldsymbol{\Psi}_{c}, \boldsymbol{\Psi}, \boldsymbol{\Phi}_{c}  \\
%	\boldsymbol{\Phi}, \mathbf{z}_i, \mathbf{u}_i,\mathbf{v}_i
%	\end{subarray}
%}{\min} 
%\,
%\sum\limits_i
%\left\| 
%\begin{bmatrix}
%\mathbf{z}_i \\
%\mathbf{u}_i \\
%\mathbf{v}_i \\
%\end{bmatrix} 
%\right\|_0
%\,
%%\\
%\textrm{s.t. }
%\, \! \!
%%&  \! \! \! \! \!
%\sum\limits_i
%\left\|
%\begin{bmatrix} 
%\mathbf{x}^{ns}_{i} \\ \mathbf{y}_{i}
%%\mathbf{x}^{ns}_{i} \\ \gamma \, \mathbf{y}_{i}
%\end{bmatrix}
%-
%\begin{bmatrix}
%\boldsymbol{\Psi}_{c} \; \boldsymbol{\Psi} \; \mathbf{0} \\
%\boldsymbol{\Phi}_{c} \; \mathbf{0} \; \boldsymbol{\Phi} \\
%\end{bmatrix}
%%
%\begin{bmatrix}
%\mathbf{z}_i \\
%\mathbf{u}_i \\
%\mathbf{v}_i \\
%\end{bmatrix}
%\right\|_2^2
%\leq C \, \sigma^2
%%\end{array}
%\end{equation}
%\textcolor{blue}{
where, $\mathbf{D}$ and $\boldsymbol{\alpha}$ represent all the dictionaries and sparse representations. The $\ell_0$ term serves as the sparsity-inducing operation.
%}
%\footnote{
%	We denote by $\| x\|_0$ the number of nonzero elements of the vector $x$. Note, this $\ell_0$ sparsity measure is a pseudo norm as it does not preserve the homogeneity property. 
%	%	It can also be relaxed to $\ell_1$-penalty. We denote by $\| x\|_1:= \sum_{i=1}^{m} |x[i]| $ the sum of absolute value of each element.
%	%	Generally, the $\ell_p$ norm of a vector $x \in \mathbb{R}^m$ is defined, for $p>1$, by $\| x \|_p := \left( \sum_{i=1}^{m} |x[i]|^p \right)^{1/p}$. Following the tradition, the $\ell_p$ pseudo norm for $p<1$ is defined by $\| x \|_p := \sum_{i=1}^{m} |x[i]|^p$.
%}
%where the standard deviation $\sigma$ represents the noise level, and $C$ is a constant. The $\ell_0$ term serves as the sparsity-inducing operation. 
The quadratic "data-fitting" term ensures that each pair of multimodal image patches is well approximated by their sparse representations with respect to the learned dictionaries. The coupled dictionary learning problems in \eqref{Eq:CDLDN} is a non-convex optimization problem. We solve it via an alternating optimization scheme that performs 1) sparse coding and 2) dictionary update alternatively. During the sparse coding stage, we fix all the dictionaries and obtain the sparse representations using orthogonal matching pursuit (OMP) algorithm~\cite{tropp2007signal}, while during the dictionary updating stage, we fix the sparse codes and update the all the dictionaries using adapted K-SVD algorithm~\cite{aharon2006img}. 
%The dictionary updating formulations are adapted from Block Coordinate Descent algorithm~\cite{mairal2010online}, where we train the common dictionaries simultaneously while train the unique dictionaries individually.
%
In our case, we stick to $\ell_0$ penalty for the sparse coding as it usually leads to better denoising performance than $\ell_1$ penalty, which is also observed in \cite{mairal2009non,mairal2014sparse}. 

\vspace{-0.2cm}

\mypar{Step 2: Reconstruction}
In this stage, given the sparse codes $\mathbf{z}_i,\mathbf{u}_i$, the denoised image can be estimated from the noisy version by solving 
%\eqref{Eq:Denoise_Update}
\begin{equation} \label{Eq:Denoise_Update}
 \footnotesize
%\{\mathbf{\hat{x}}_{}, \mathbf{z},\mathbf{u},\mathbf{v} \} = 
\underset{\mathbf{X}_{} }{\min} \;
\mu \left\| \mathbf{X}_{} - \mathbf{X}^{ns}_{} \right\|_2^2
+
\sum_i
\left\| \mathbf{R}_i \mathbf{X} - ( \boldsymbol{\Psi}_{c} \mathbf{z}_i + \boldsymbol{\Psi} \mathbf{u}_i) \right\|_2^2
\end{equation}
%\textcolor{blue}{
	where, $\mu$ trades off between the fidelity to the noisy version and the fidelity to the sparse estimation.\footnote{For $\mu = 0$, this expression just represents the average of the denoised image patches on the overlapping areas, leading to a purified image. 
%		For non-zero $\mu$, the estimation operation also refers to the original noisy image during the average.
%}
}
This leads immediate to a closed form solution
%\eqref{Eq:Denoise_Update2}.
%% penalty version 
%
\begin{equation} \label{Eq:Denoise_Update2}
%\small
\footnotesize
\mathbf{X} = \Big( \mu \mathbf{I} + \sum_{i} \mathbf{R}_i^T \mathbf{R}_i \Big)^{-1} 
\Big( \mu \mathbf{X}^{ns} + \sum_i \mathbf{R}_i^T (\boldsymbol{\Psi}_{c} \mathbf{z}_i + \boldsymbol{\Psi} \mathbf{u}_i) \Big) \,.
\end{equation} 
%The coupled image denoising algorithm is described in Algorithm~\ref{Alg:CoupledDenoise}. 

%\vspace{-0.2cm}

\section{Multimodal Image Super-resolution}
\label{sec:SIMIS}

\vspace{-0.2cm}

We now present our coupled dictionary learning framework for multi-modal image super-resolution.

\vspace{-0.3cm}

\subsection{Data Model for Super-resolution}
%\textcolor{blue}{
	Similar to section 2.1, we consider a pair of vectorized low-resolution (LR) and high-resolution (HR) images $\mathbf{X}^l \in \mathbb{R}^{M}$ and $\mathbf{X}^h \in \mathbb{R}^{N}$ of one modality and a corresponding registered HR guidance image $\mathbf{Y} \in \mathbb{R}^{N}$ of different modality as side information. In particular, let $\mathbf{x}^{l}_{i} \in \mathbb{R}^{m}$ and $\mathbf{x}^{h}_{i} \in \mathbb{R}^{n}$ denote the $i$-th LR/HR image patch pair extracted from $\mathbf{X}^{l}$ and $\mathbf{X}^{h}$ and let $\mathbf{y}_{i} \in \mathbb{R}^{n}$ denote the corresponding $i$-th HR guidance image patch extracted from $\mathbf{Y}$. Then, based on the following two main assumptions: (1) $\mathbf{x}^{l}_i$ and $\mathbf{x}^{h}_i$ of the same modality share the same sparse representations with respect to their own dictionaries, which is also adopted by \cite{yang2008image,yang2010image,zeyde2010single}; (2) a pair of registered image patches from different modalities $\mathbf{x}^{h}_i$ and $\mathbf{y}_i$ admit both identical and diverse sparse representations, we propose the data model \eqref{Eq:SparseModelX} - \eqref{Eq:SparseModelY} to capture the relationship across the LR/HR patches of a certain image modality along with the corresponding patch of another image modality. 
%}
%
\begin{align}
\mathbf{x}^{h}_i &= \boldsymbol{\Psi}_{c}^{h} \, \mathbf{z}_i + \boldsymbol{\Psi}^{h} \, \mathbf{u}_i \,,
\label{Eq:SparseModelX}
\\
\mathbf{x}^{l}_i 
&= \boldsymbol{\Psi}_{c}^{l} \, \mathbf{z}_i + \boldsymbol{\Psi}^{l} \, \mathbf{u}_i \,,
\label{Eq:SparseModelX_low}
\\
\mathbf{y}_i &= \boldsymbol{\Phi}_{c} \, \mathbf{z}_i + \boldsymbol{\Phi} \, \mathbf{v}_i \,,
\label{Eq:SparseModelY}
\end{align}
%%
%\noindent where $\boldsymbol{\Psi}_{c}^{l} = \mathbf{A} \boldsymbol{\Psi}_{c}^{h}$ and $ \boldsymbol{\Psi}^{l} = \mathbf{A} \boldsymbol{\Psi}^{h}$. 
where, similar to \eqref{Eq:SparseModelX_Noise} and \eqref{Eq:SparseModelY_Noise}, $\mathbf{z}_i  \in \mathbb{R}^{K}$ is the common sparse representation, while $\mathbf{u}_i  \in \mathbb{R}^{K}$ and $\mathbf{v}_i  \in \mathbb{R}^{K}$ are the unique sparse representations. $\boldsymbol{\Psi}_{c}^{h} $, $\boldsymbol{\Psi}_{c}^{l}$ and $\boldsymbol{\Phi}_{c}  \in \mathbb{R}^{n \times K}$ are the dictionaries associated with $\mathbf{z}_i$, whereas $\boldsymbol{\Psi}^{h}$, $\boldsymbol{\Psi}^{l} $ and $\boldsymbol{\Phi} \in \mathbb{R}^{n \times K}$ are dictionaries associated with $\mathbf{u}$ and $\mathbf{v}$, respectively.
Note that the sparse vectors $\mathbf{z}_i$ and $\mathbf{u}_i$ capture the relationship between the LR and HR patches of the same modality in \eqref{Eq:SparseModelX} and \eqref{Eq:SparseModelX_low}. Moreover, the common sparse vector $\mathbf{z}_i$ connects the various patches of the two different modalities in \eqref{Eq:SparseModelX} - \eqref{Eq:SparseModelY}. The disparities between modalities are distinguished by the sparse vectors $\mathbf{u}_i$ and $\mathbf{v}_i$. 

%Overall, this data model allows each pair of patches to be non-linearly transformed to a sparse domain where it is allowed to characterize the similarities and disparities between different modalities.

%
%By capitalizing on this model, we propose in the sequel a novel joint image SR scheme that consists of two stages: (1) a training stage referred to as coupled dictionary learning (CDL) and (2) a testing stage referred to as coupled image super-resolution (CSR) (see Figure~\ref{Fig:Diagram}). In the training stage, we learn the dictionaries in \eqref{Eq:SparseModelX} - \eqref{Eq:SparseModelY} from a set of training image patches to couple different data modalities together. Then, in the testing stage, we use the learned dictionaries to find the representations of the LR testing patch and corresponding HR guidance patch, according to \eqref{Eq:SparseModelX_low} and \eqref{Eq:SparseModelY}. These sparse representations are then used to reconstruct the desired HR target image patch via \eqref{Eq:SparseModelX}.

\vspace{-0.2cm}

\subsection{Coupled Super Resolution (CSR)}
Our coupled image super-resolution problem is addressed in 3 steps: coupled dictionary learning, coupled sparse coding and reconstruction of the HR image.

\mypar{Step 1: Coupled dictionary learning}
Given a set of training patch pairs $\{\mathbf{x}^{l}_i, \mathbf{x}^{h}_i, \mathbf{y}_i \}$, we solve the following optimization problems \eqref{Eq:CDL1} to train a group of dictionaries.
%
%% L0 norm of the sparse codes as the constraints
\begin{equation} \label{Eq:CDL1}
\footnotesize
%\begin{array}{cl}
\underset{ \mathbf{D}, \boldsymbol{\alpha}
}{\text{min}} 
%& % ! \! \! \!
\;
\sum\limits_i
\left\|
\begin{bmatrix} 
\mathbf{x}^{h}_i \\ \mathbf{x}^{l}_i \\ \mathbf{y}_i
\end{bmatrix}
-
\begin{bmatrix}
\boldsymbol{\Psi}_{c}^h \; \boldsymbol{\Psi}^h \; \mathbf{0} \\
\boldsymbol{\Psi}_{c}^l \;\, \boldsymbol{\Psi}^l \;\, \mathbf{0} \\
\boldsymbol{\Phi}_{c} \;\,\, \mathbf{0} \;\;\; \boldsymbol{\Phi} \\
\end{bmatrix}
\begin{bmatrix}
\mathbf{z}_i \\
\mathbf{u}_i \\
\mathbf{v}_i \\
\end{bmatrix}
\right\|_2^2
%\\
\;\;
\text{s.t.}
\;
%& % \! \! \! \!
\left\| 
\begin{bmatrix}
\mathbf{z}_i \\
\mathbf{u}_i \\
\mathbf{v}_i \\
\end{bmatrix} 
\right\|_0
\leq s_i
\,, \forall i,
%\end{array}
\end{equation}

\noindent
where the $\ell_2$ norm promotes the data fidelity; the $\ell_0$ pseudo-norm promotes sparsity for the sparse codes; $\mathbf{D}$ and $\boldsymbol{\alpha}$ represent all the dictionaries and sparse representations.

%and the Frobenius norm is to avoid extremely large value for corresponding atoms, making the solution more stable.
%
%Compared with conventional sparse coding problems that involves only LR image patch $\mathbf{x}^{l}$, our formulations~\eqref{Eq:SR} also integrates the side information $\mathbf{y}_{test}$ into the sparse coding task. Since the increase in the amount of available information is akin to the increase of the number of measurements in a Compressive Sensing scenario~\cite{mota2017compressed,renna2016classification}, one can expect to obtain a more accurate estimate of the sparse codes. 

\mypar{Step 2: Coupled sparse coding}
Given a new LR testing image and a corresponding registered HR guidance image as side information. We extract (overlapping) image patch pairs from these two modalities. 
%\textcolor{blue}{
Then we solve a sparse coding problem, similar to \eqref{Eq:CDL1} but with fixed dictionaries learned from Step 1 and involving no $\mathbf{x}^{h}_i$ and $[\boldsymbol{\Psi}_c^h, \boldsymbol{\Psi}^h]$.
%}

%% image No7 -- denoised images for only 3 noise levels
\begin{figure}[t]
	\centering
	%---------------------------------------------------
	% im No. 7
	\begin{minipage}[b]{0.24\linewidth}
		\centering
		{\footnotesize urban\_0030} \hfill % \vfill
		\includegraphics[width = 2cm, height = 2cm]{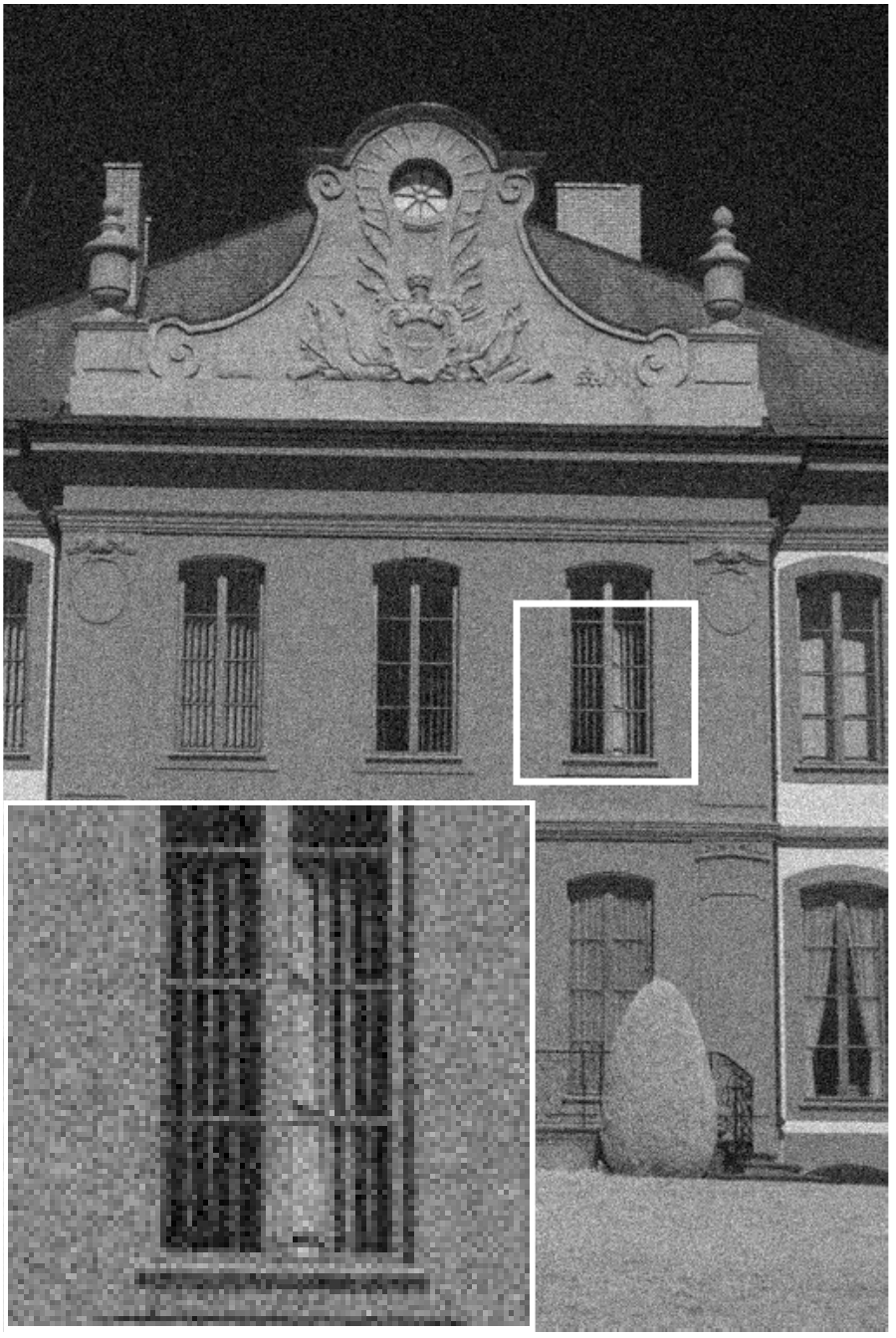}
	\end{minipage} 
	\begin{minipage}[b]{0.24\linewidth}
		\centering
		\includegraphics[width = 2cm, height = 2cm]{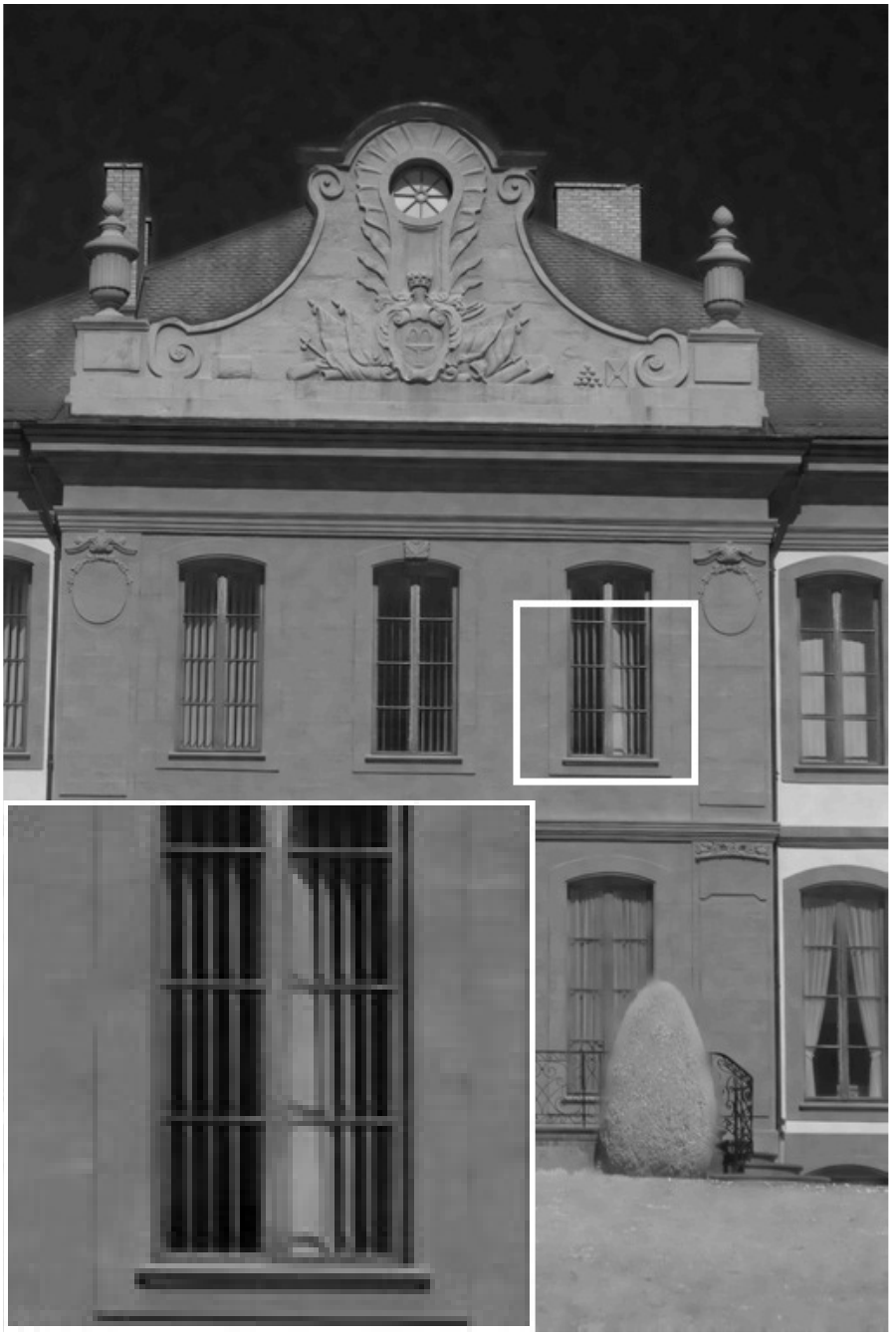}
	\end{minipage} 
	\begin{minipage}[b]{0.24\linewidth}
		\centering
		\includegraphics[width = 2cm, height = 2cm]{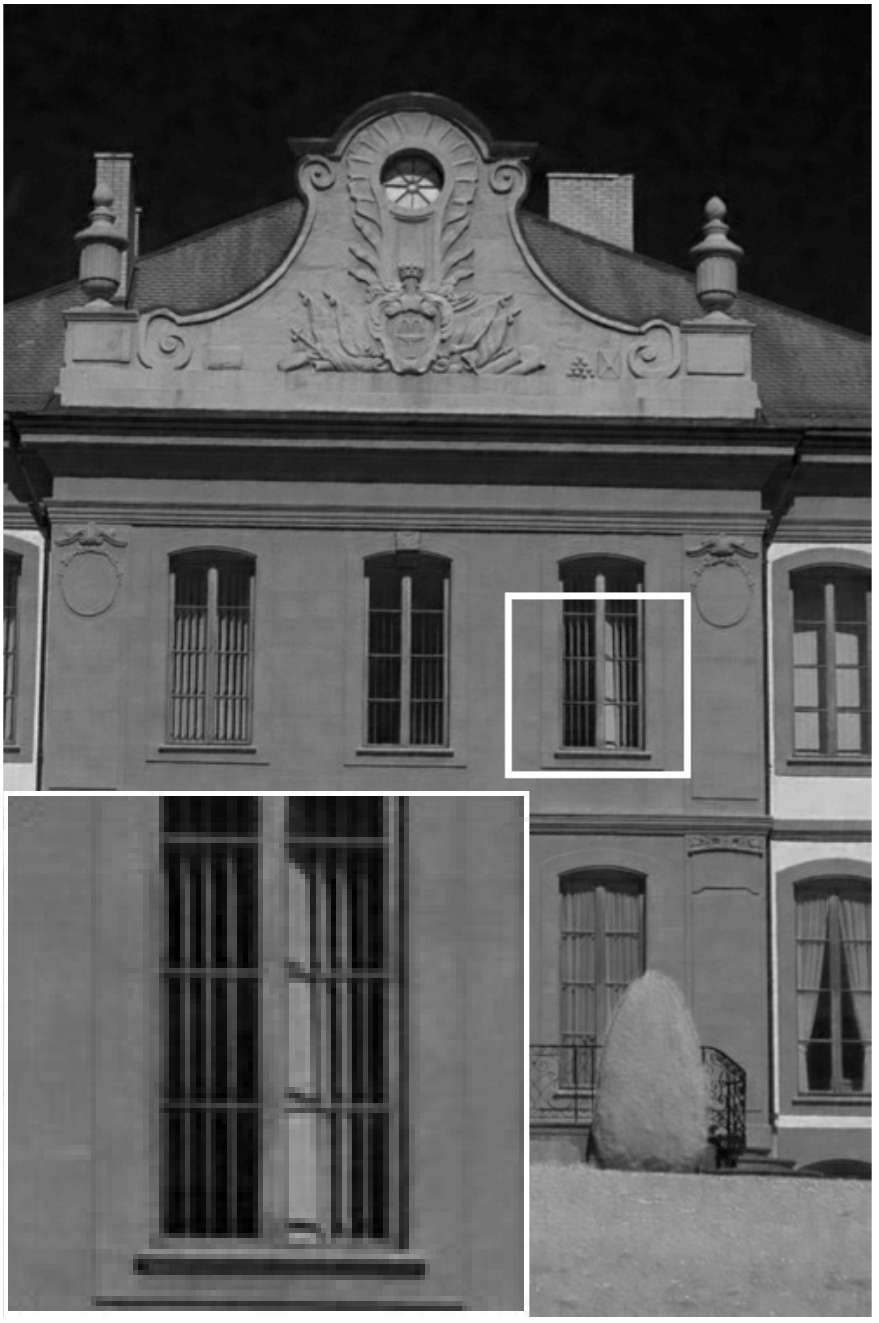}
	\end{minipage} 
	\begin{minipage}[b]{0.24\linewidth}
		\centering
		\includegraphics[width = 2cm, height = 2cm]{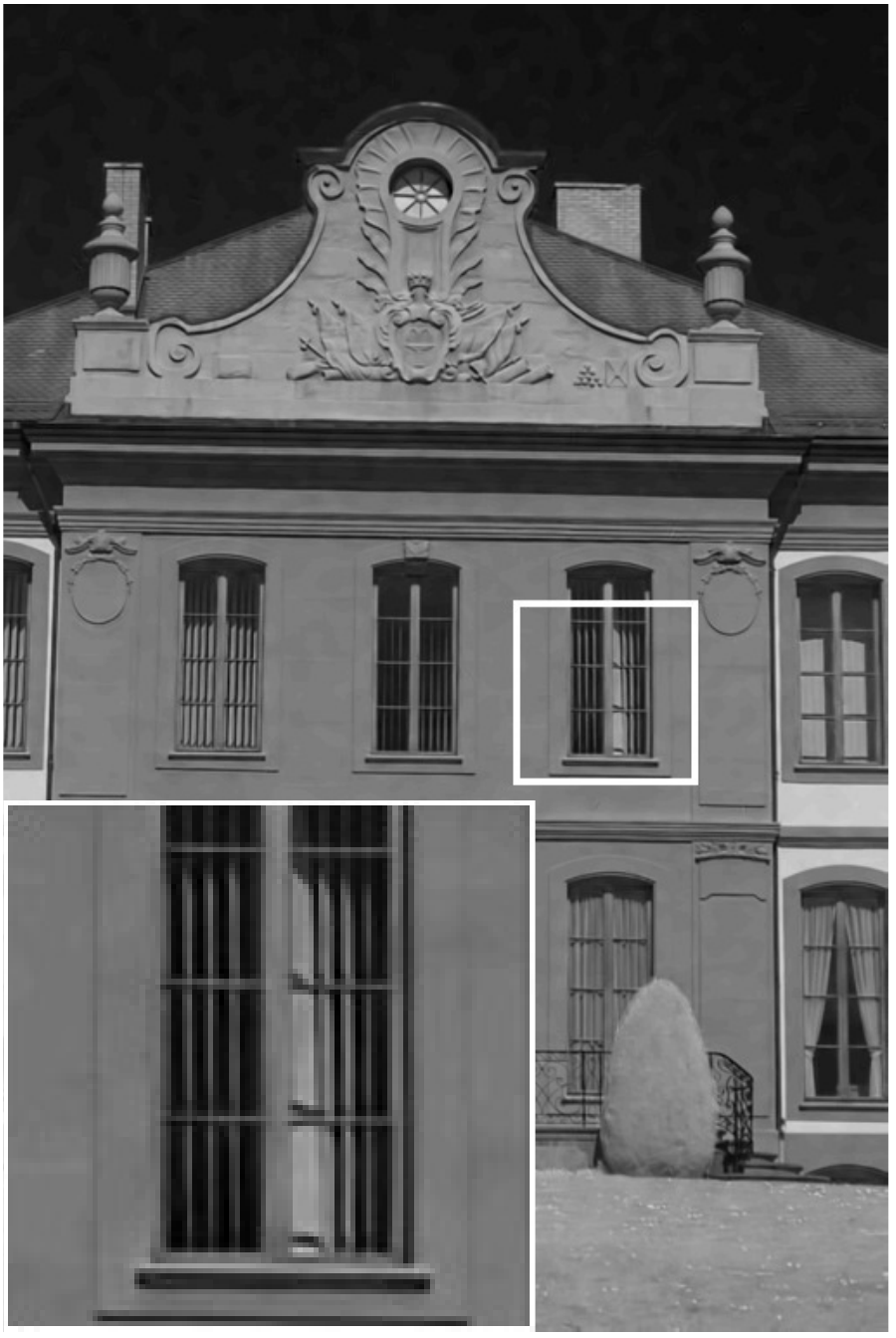}
	\end{minipage} 
	\\
	% -------------------------------------------
	% SI and residue
	\begin{minipage}[b]{0.24\linewidth}
		\centering
		\includegraphics[width = 2cm, height = 2cm]{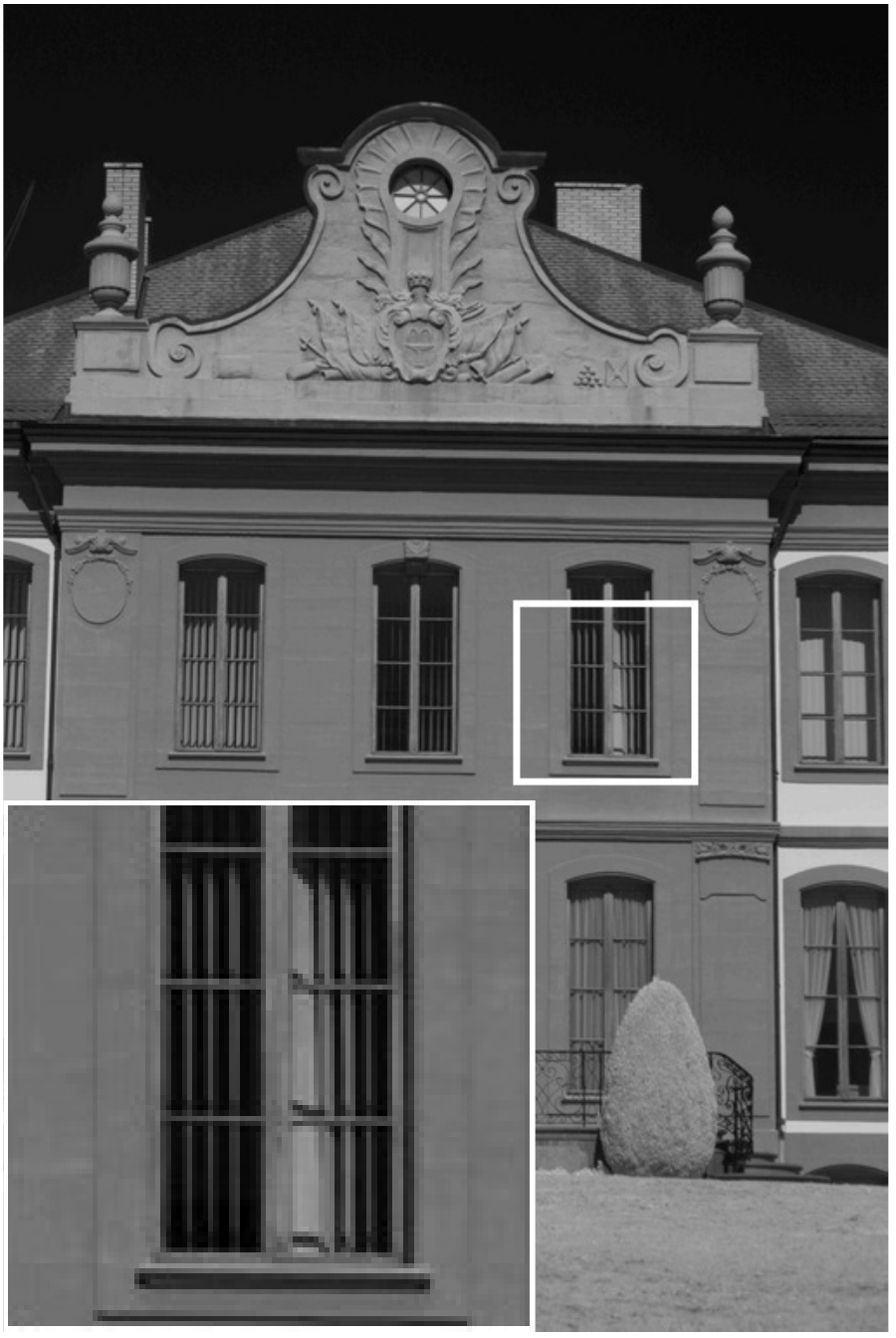}
	\end{minipage} 
	\begin{minipage}[b]{0.24\linewidth}
		\centering
		\includegraphics[width = 2cm, height = 2cm]{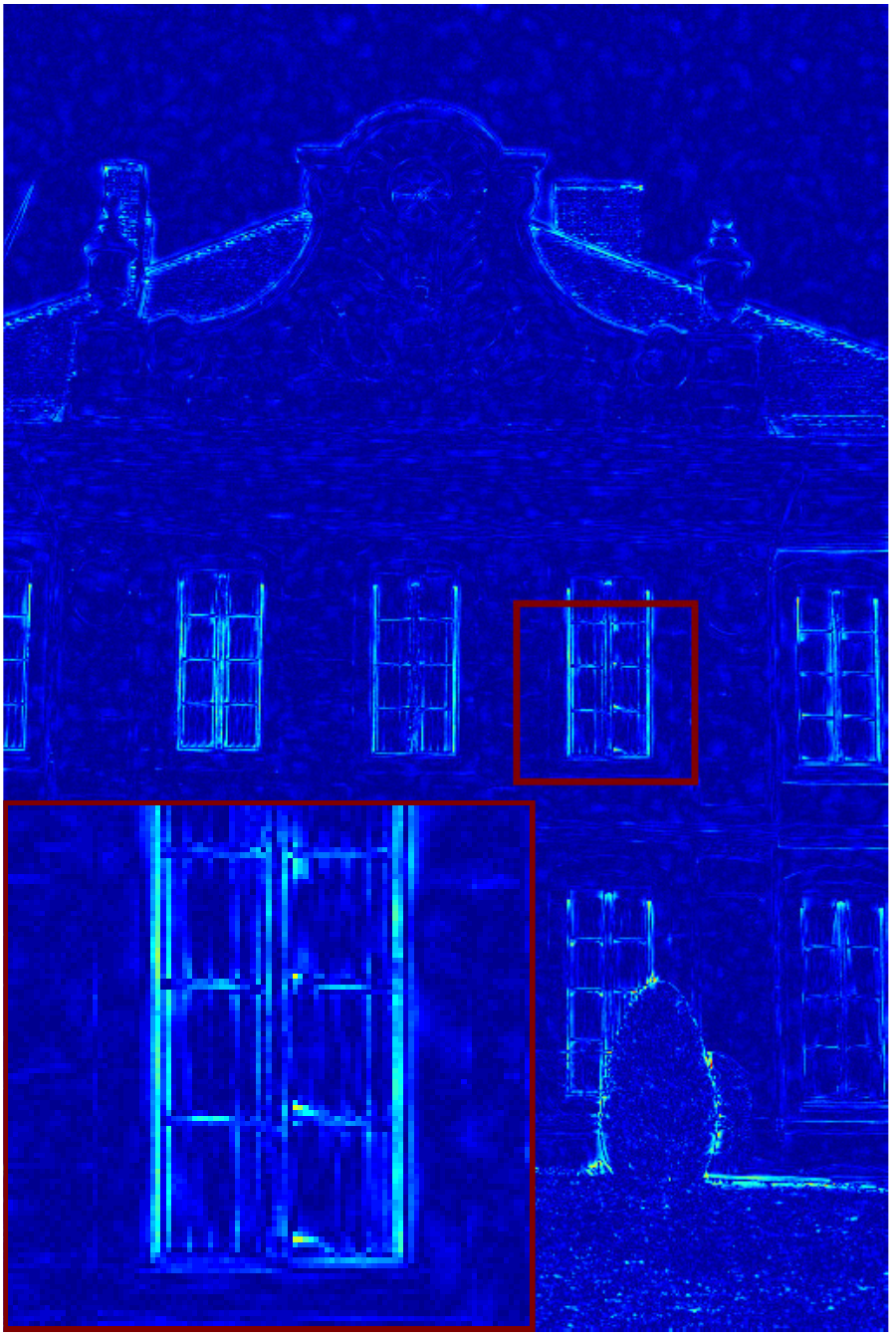}
	\end{minipage} 
	\begin{minipage}[b]{0.24\linewidth}
		\centering
		\includegraphics[width = 2cm, height = 2cm]{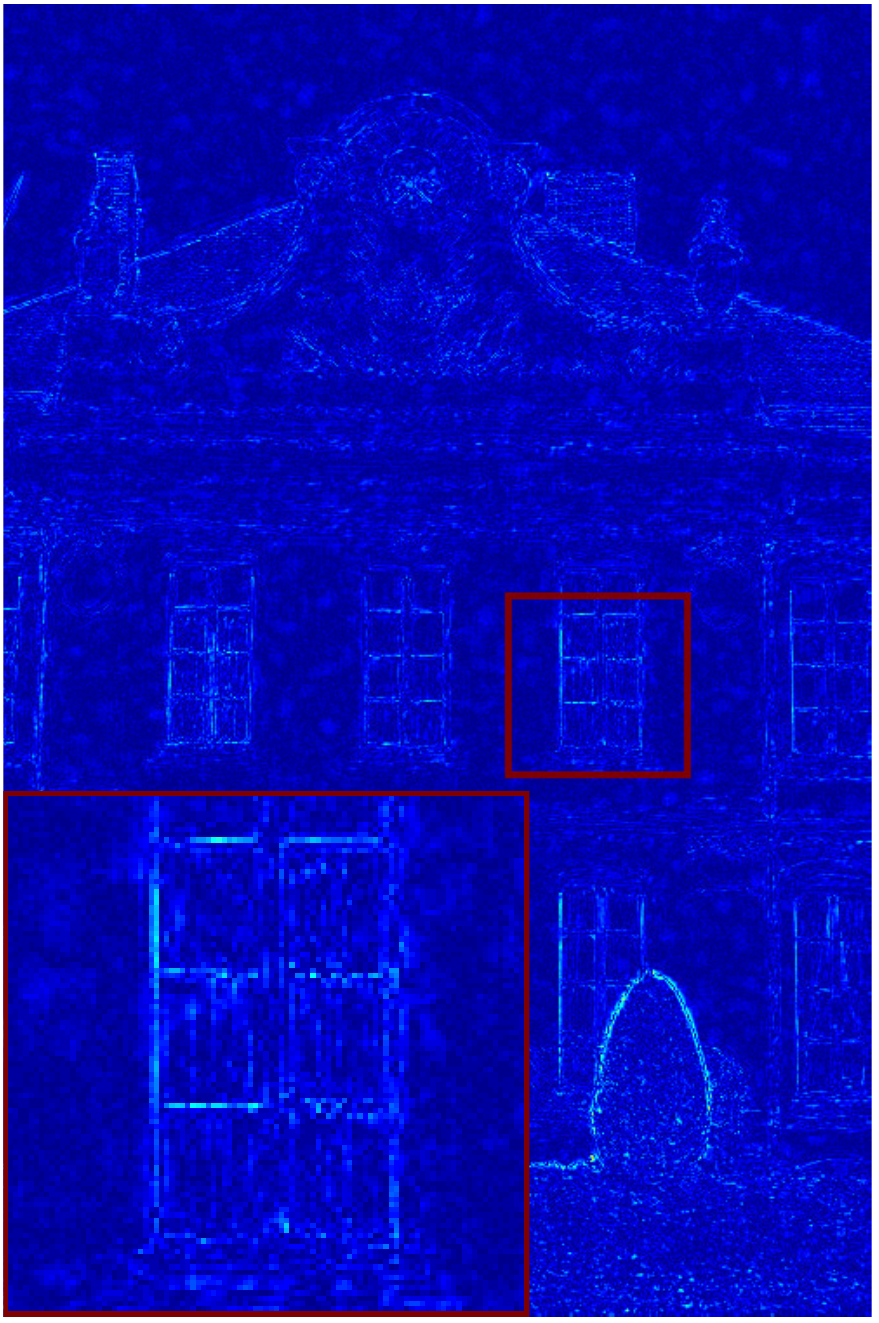}
	\end{minipage} 
	\begin{minipage}[b]{0.24\linewidth}
		\centering
		\includegraphics[width = 2cm, height = 2cm]{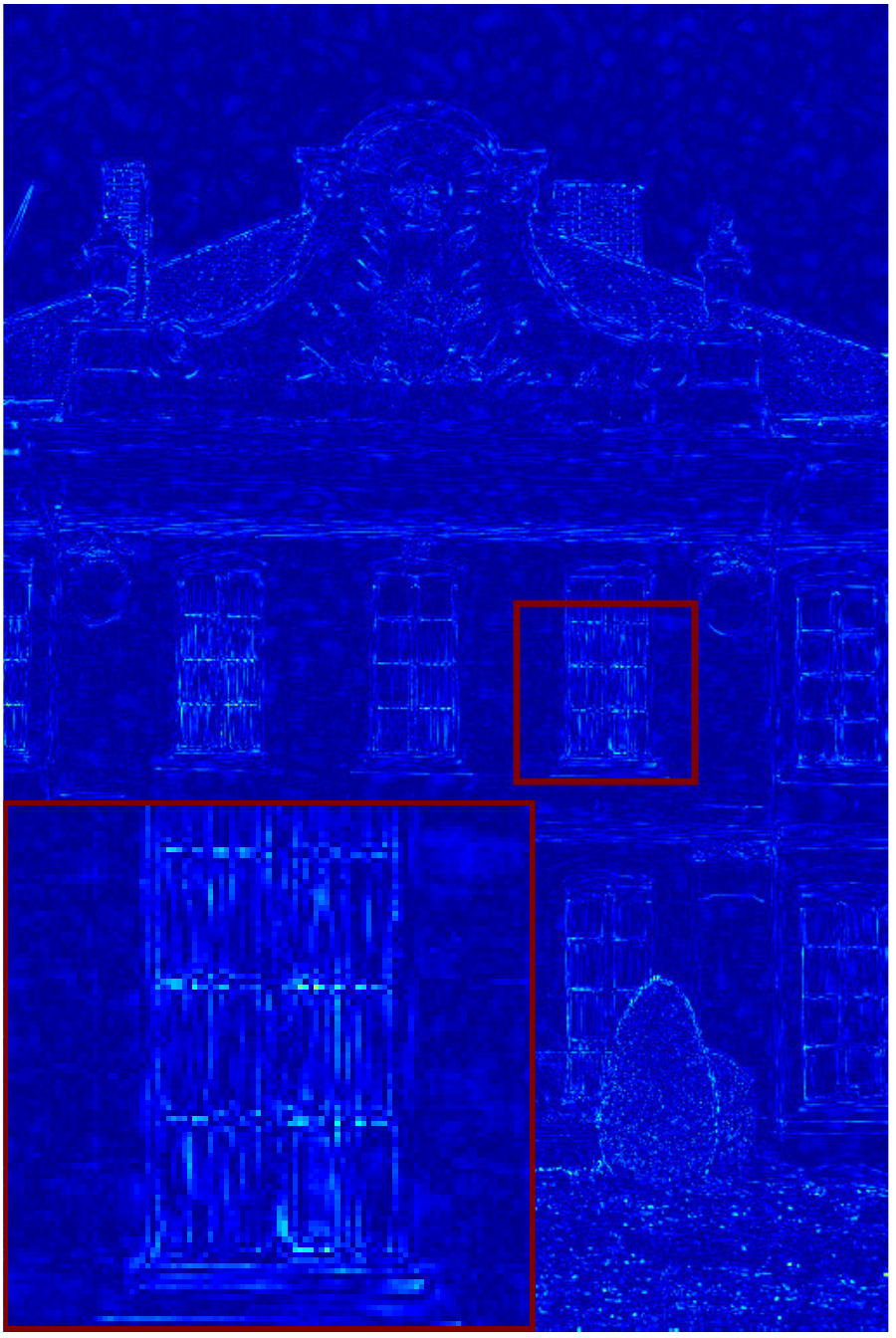}
	\end{minipage} 
	\\
	%---------------------------------------------------
	% im No. 8
	\begin{minipage}[b]{0.24\linewidth}
		\centering
		{\footnotesize urban\_0050} \hfill % \vfill
		\includegraphics[width = 2cm, height = 2cm]{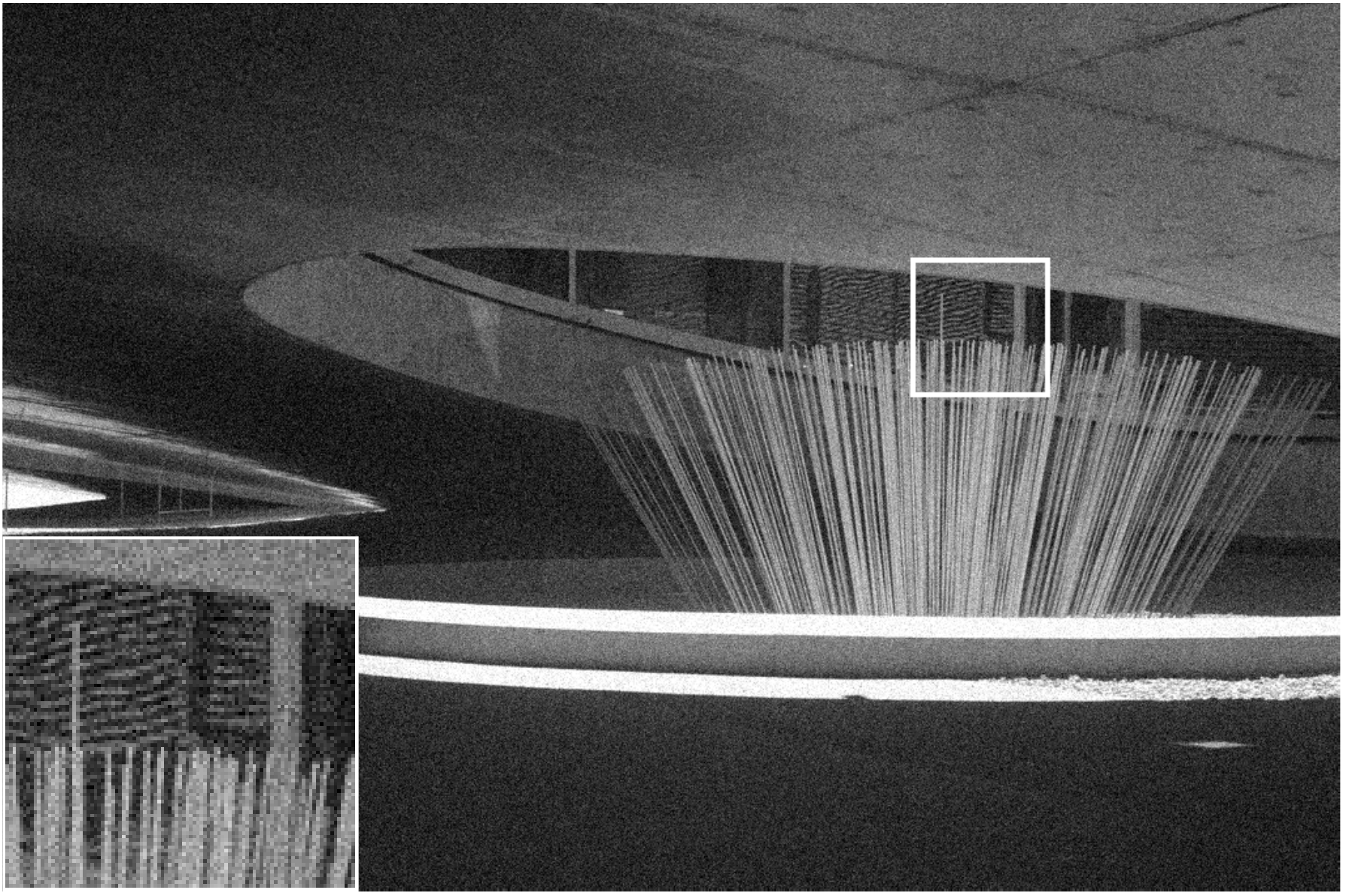}
	\end{minipage} 
	\begin{minipage}[b]{0.24\linewidth}
		\centering
		\includegraphics[width = 2cm, height = 2cm]{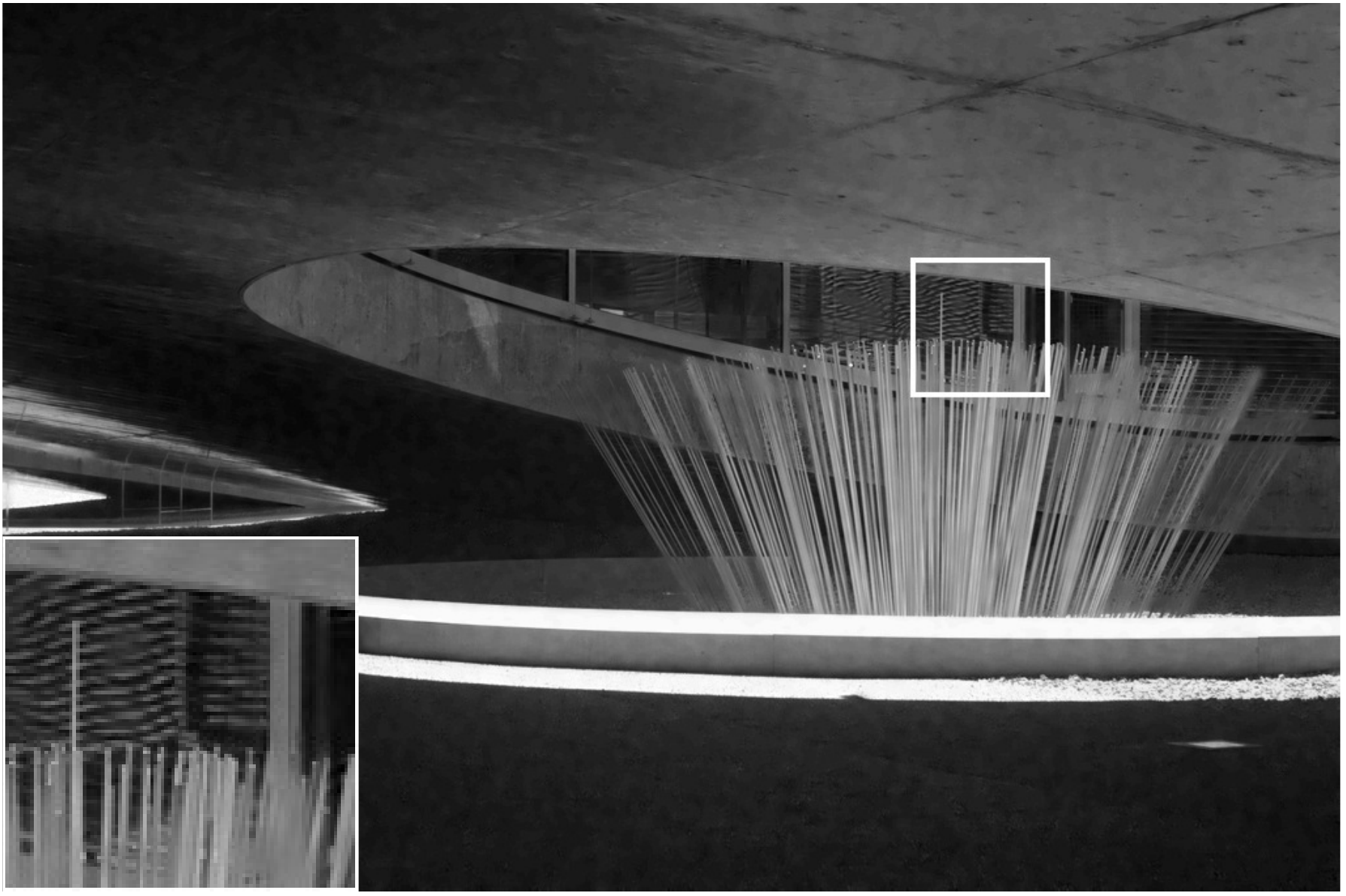}
	\end{minipage} 
	\begin{minipage}[b]{0.24\linewidth}
		\centering
		\includegraphics[width = 2cm, height = 2cm]{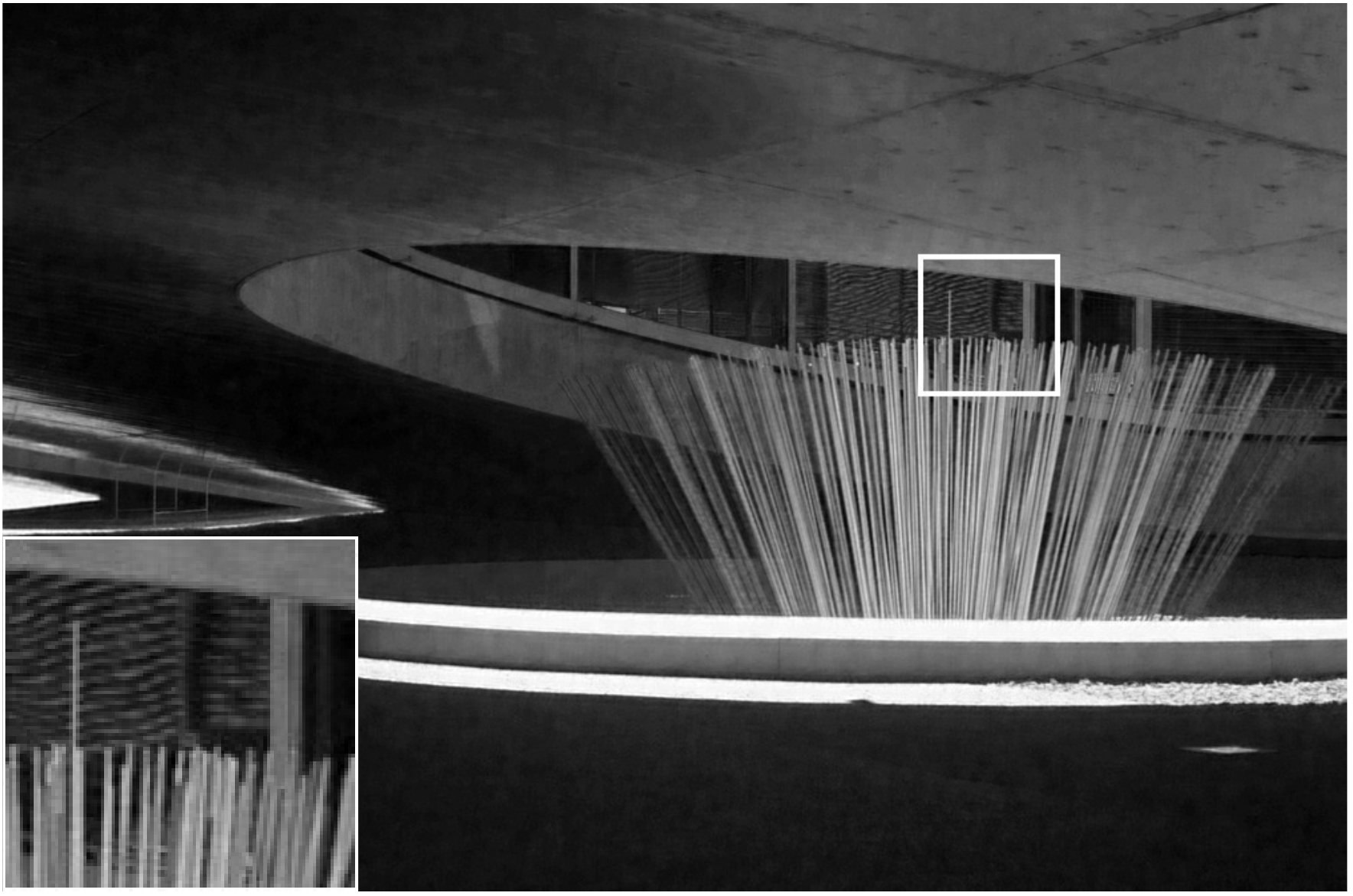}
	\end{minipage} 
	\begin{minipage}[b]{0.24\linewidth}
		\centering
		\includegraphics[width = 2cm, height = 2cm]{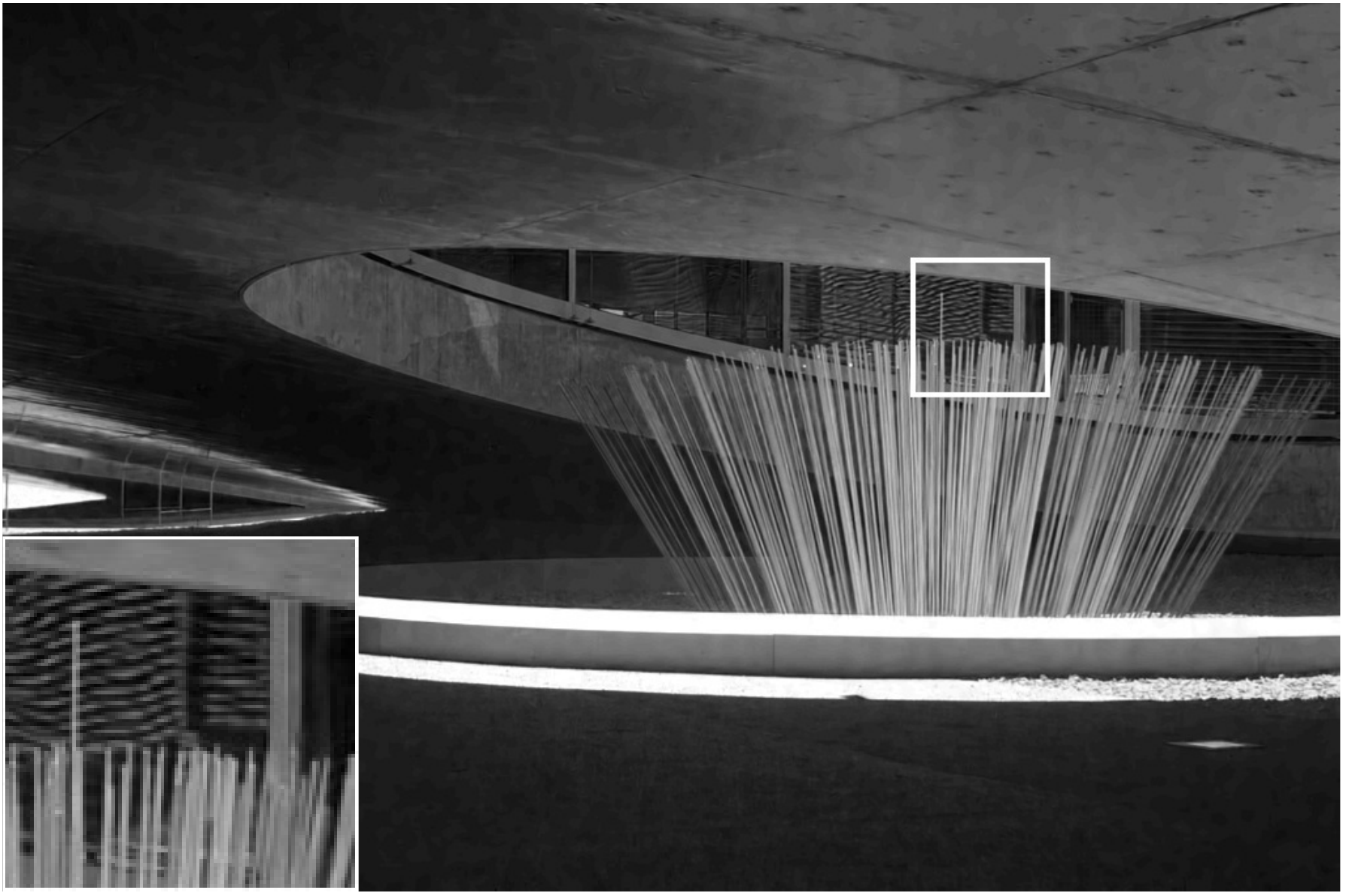}
	\end{minipage} 
	\\
	% -------------------------------------------
	% SI and residue
	\begin{minipage}[b]{0.24\linewidth}
		\centering
		\includegraphics[width = 2cm, height = 2cm]{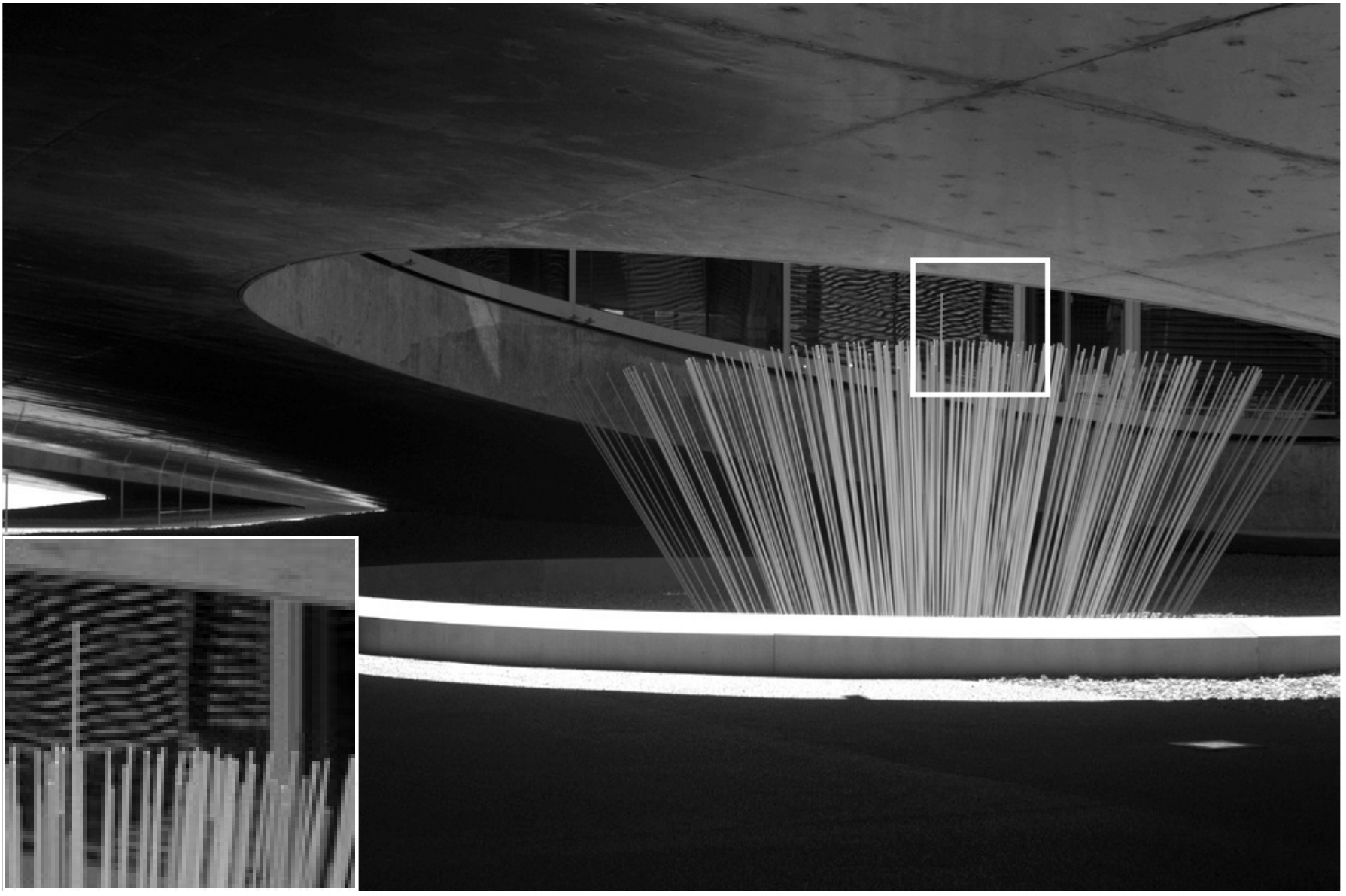}
	\end{minipage} 
	\begin{minipage}[b]{0.24\linewidth}
		\centering
		\includegraphics[width = 2cm, height = 2cm]{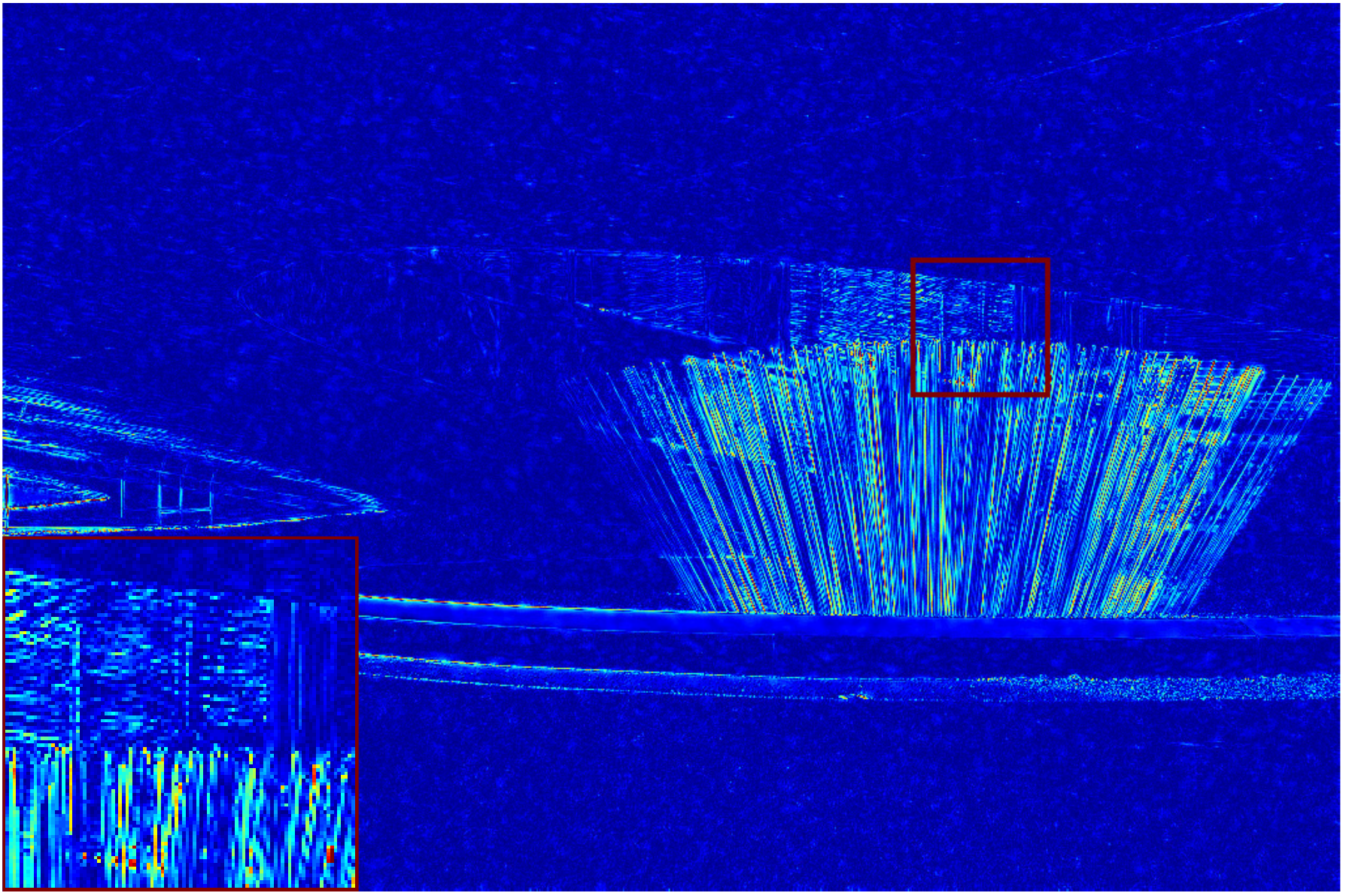}
	\end{minipage} 
	\begin{minipage}[b]{0.24\linewidth}
		\centering
		\includegraphics[width = 2cm, height = 2cm]{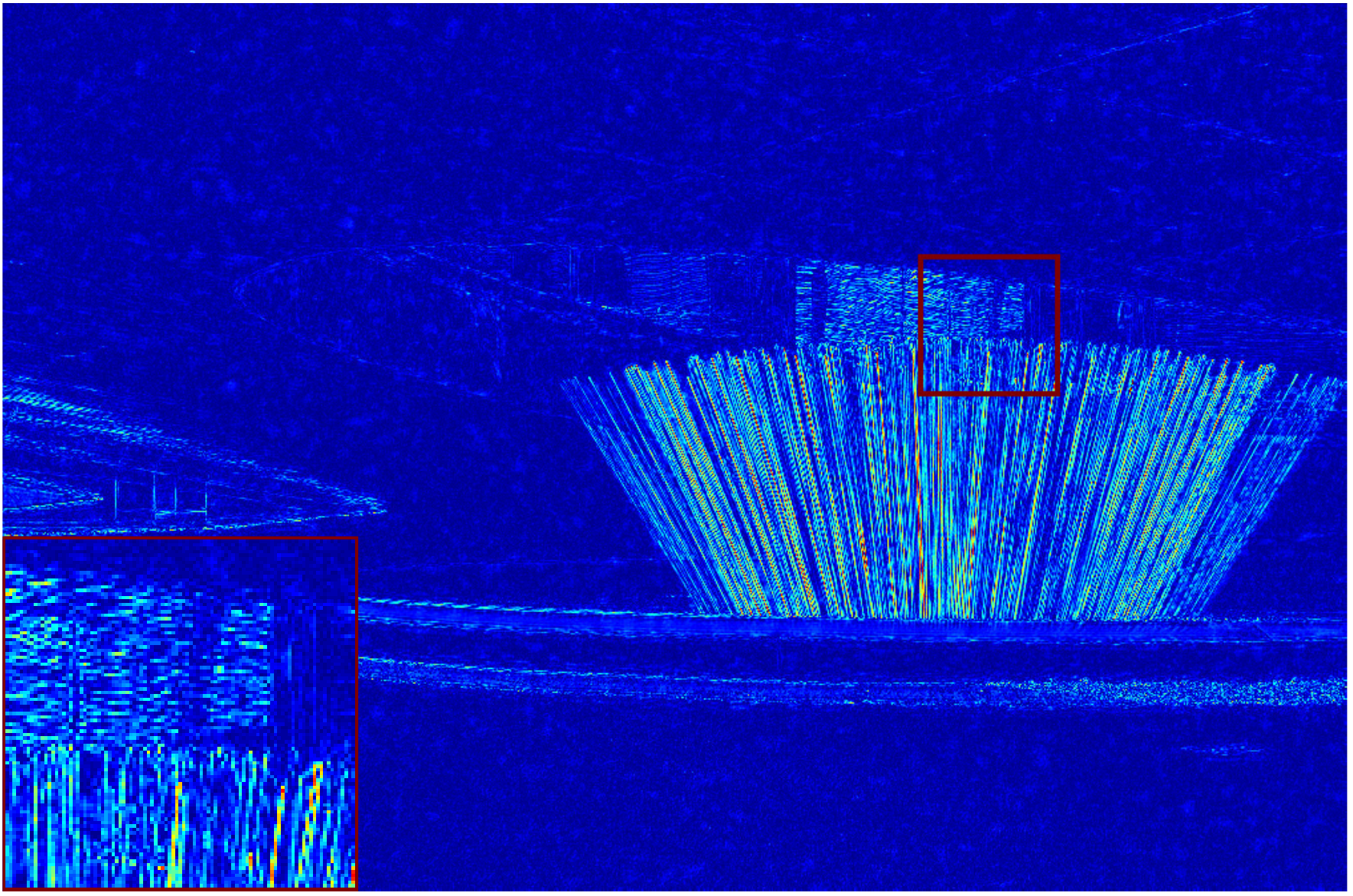}
	\end{minipage} 
	\begin{minipage}[b]{0.24\linewidth}
		\centering
		\includegraphics[width = 2cm, height = 2cm]{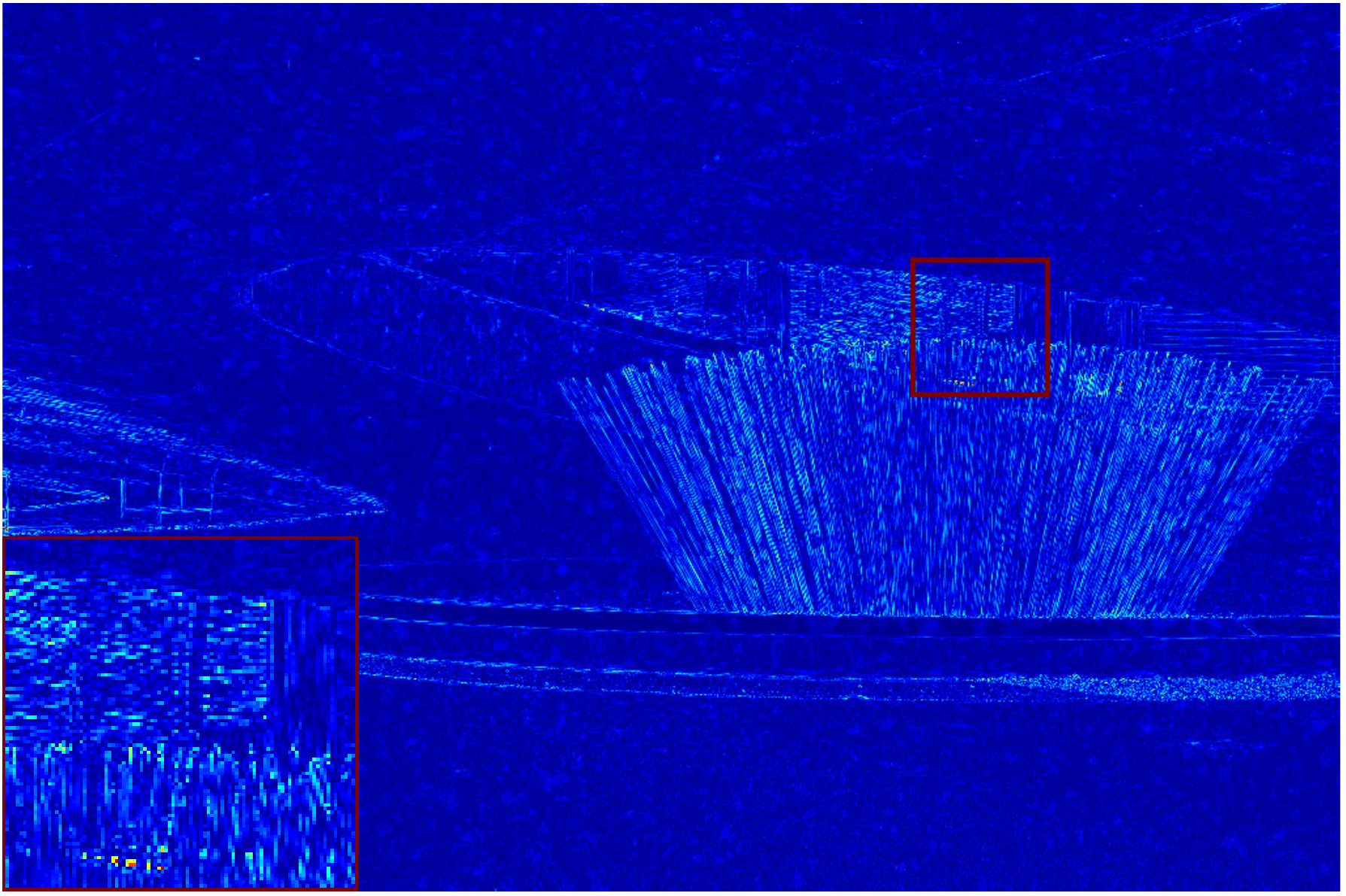}
	\end{minipage} 
	\\
	%---------------------------------------------------
	%	\begin{minipage}[b]{0.08\linewidth}
	%		\centering	{\footnotesize (a)LR}
	%	\end{minipage} 
	\begin{minipage}[b]{0.24\linewidth}
		\centering	{\footnotesize (a) Input/Truth} 
	\end{minipage}
	\begin{minipage}[b]{0.24\linewidth}
		\centering	{\footnotesize (b) GF~\cite{he2013guided}} 
	\end{minipage} 
	\begin{minipage}[b]{0.24\linewidth}
		\centering	{\footnotesize (c) DJF\cite{li2016deep}} 
	\end{minipage} 
	\begin{minipage}[b]{0.24\linewidth}
		\centering 	{\footnotesize (d) Proposed}
	\end{minipage} 
	%----------------------------------------------------------
	
	\vspace{-0.3cm}
	
	\begin{multicols}{1}  
		\begin{minipage}[b]{0.98\linewidth}
			%			{\footnotesize Colorbar}
			\includegraphics[width = 8.5cm, height=0.3cm]{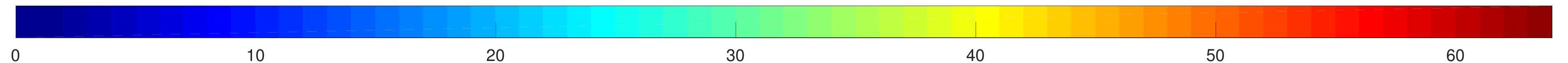}
		\end{minipage}
	\end{multicols}	
	
	\vspace{-0.8cm}
	
	%----------------------------------------------------------
	\caption{Multimodal image denoising for near-infrared images ($\sigma=16$). For each scene, the first row is the noisy input and denoised results and the second row is the ground truth and corresponding error map for each approach.}
	%		 In the error map, brighter area represents larger error.}
	\label{Fig:DenoisedIms_7}
\end{figure}
%
% show PSNR and RMSE
\begin{figure}[thbp]
	\centering
	\begin{minipage}[b]{0.48\linewidth}
		\centering
		\includegraphics[width = 4.4cm, height = 2.8cm]{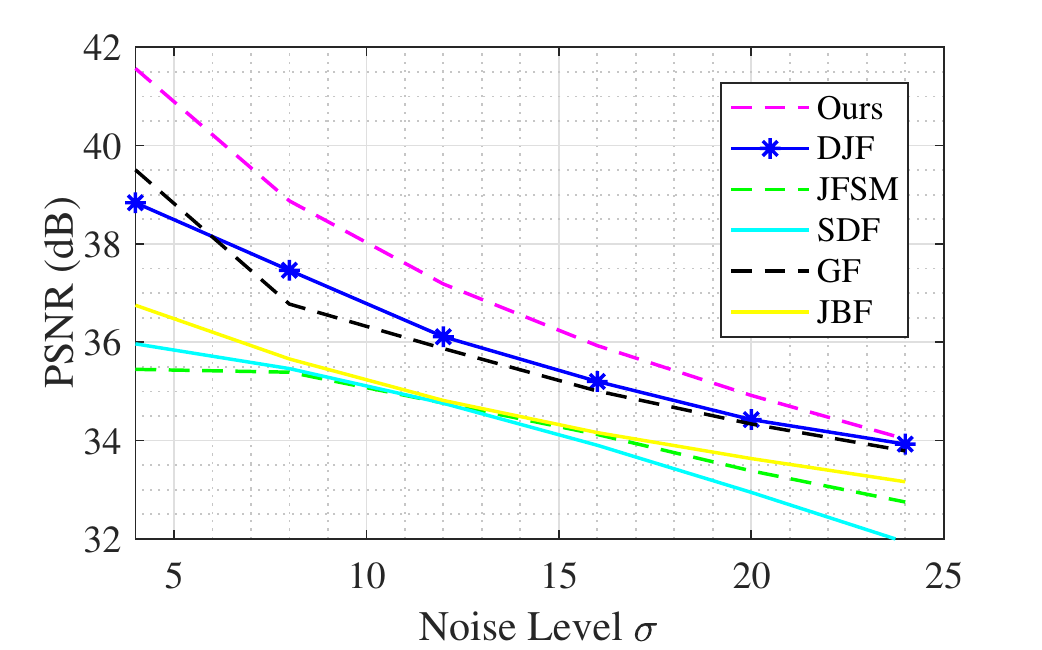} 
	\end{minipage}
	\begin{minipage}[b]{0.48\linewidth}
		\centering
		\includegraphics[width = 4.4cm, height = 2.8cm]{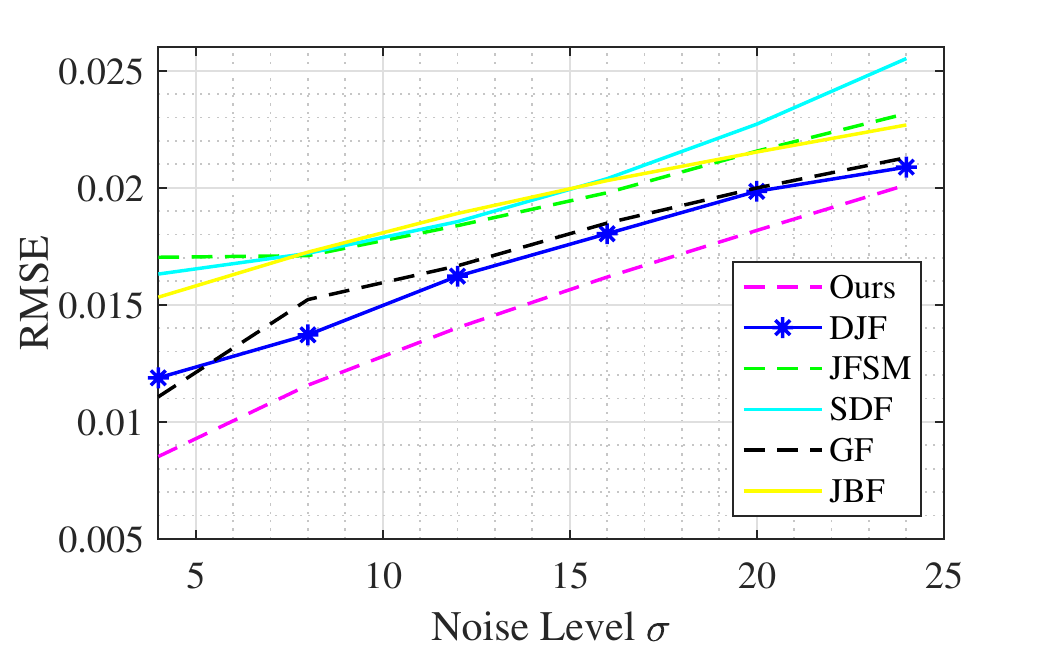}
	\end{minipage}
	
	\vspace{-0.2cm}
	
	\caption{Multimodal image denoising in terms of PSNR and RMSE with respect to different noise levels. We compare our approach with state-of-the-art joint image filtering approaches, JBF\cite{kopf2007joint}, GF\cite{he2013guided}, SDF\cite{ham2017robust}, JFSM\cite{shen2015multispectral} and DJF\cite{li2016deep}.}
	\label{Fig:PSNR_RMSE}
\end{figure}

\mypar{Step 3: Reconstruction}
Finally, we can obtain each estimated HR patch $\mathbf{x}^{h}_i$ of the target image from the HR dictionaries $[\boldsymbol{\Psi}_c^h, \boldsymbol{\Psi}^h]$ and sparse codes $\mathbf{z}_i$ and $\mathbf{u}_i$ as follows

%Since the side information increases the number of rows, i.e., the total number of measurements, and the over-completeness of the whole dictionary, it is more likely to achieve better estimation of the sparse codes, according to the Compressive Sensing theory. As a consequence, this leads to more accurate approximation of $\mathbf{x}^{h}_{test}$ via multiplying the HR dictionaries $[\boldsymbol{\Psi}_c^h, \boldsymbol{\Psi}^h]$ with sparse representations $\mathbf{z}$ and $\mathbf{u}$, show as 
%%% analysis from the perspective of compressive sensing.
%%From the perspective of Compressive Sensing, the conventional sparse coding problem require $m \geq O(s \ln N/s)$ measurements to recovery the corresponding sparse codes, but now our formulation require $m \geq O(s \ln 3N/s)-N$ measurements to recovery $\mathbf{z},\mathbf{u}$, and $\mathbf{v}$. So, our approach requires less measurements than conventional sparse coding methods for successful reconstruction of the sparse codes. Equivalently, with the same number of measurements, our approach can lead to better reconstruction. 
%
\begin{equation} \label{Eq:SR_2}
\mathbf{x}^{h}_i = \boldsymbol{\Psi}_{c}^{h} \mathbf{z}_i + \boldsymbol{\Psi}^{h} \mathbf{u}_i \,.
\end{equation}

\noindent
Once all the HR patches are recovered, they are integrated into a whole image by averaging on the overlapping areas. 
%The coupled super-resolution algorithm is described in Algorithm~\ref{Alg:CSR}.

\vspace{-0.2cm}

\section{Multimodal Image Inpainting}
\label{sec:Inpainting}

\vspace{-0.2cm}

We finally introduce our framework for the multi-modal image inpainting task, a scenario where one needs to inpaint images with some small holes resulting from missing pixels. To deal with unobserved data, we introduce a binary mask $\mathbf{M}_i$ for $i$-th patch, which is defined as a diagonal matrix whose value on the $j$-th entry of the diagonal is 1 if the pixel $\mathbf{x}^{h}_i[j]$ is observed and 0 otherwise, similar to \cite{mairal2014sparse}. Then, we relate each corrupted image patch $\mathbf{x}^{l}_i$ with the original image patch $\mathbf{x}^{h}_i$ by a mask $\mathbf{M}_i$, as $\mathbf{x}^{l}_i = \mathbf{M}_i \mathbf{x}^{h}_i $. Similarly, there also exist dictionaries $\boldsymbol{\Psi}_{ci}^{l}= \mathbf{M}_i \boldsymbol{\Psi}_{c}^{h}$, $ \boldsymbol{\Psi}^{l}_i = \mathbf{M}_i \boldsymbol{\Psi}^{h}_i$ for the $i$-th corrupted patch. Thus, the inpainting data model is identical to the super-resolution model in \eqref{Eq:SparseModelX} - \eqref{Eq:SparseModelY}. 
%\textcolor{blue}{
We can still use \eqref{Eq:CDL1} - \eqref{Eq:SR_2} to solve the multi-modal inpainting problem, except no need to learn $[\boldsymbol{\Psi}_c^l, \boldsymbol{\Psi}^l]$ from $\mathbf{x}^{l}_i$ (since $\mathbf{M}_i$ is known).
\begin{table*}[t]
	\centering
	\scriptsize
	\caption{Multimodal image super-resolution for near-infrared images (4$\times$ upscaling)}
	\begin{tabular}{l| ll| ll| ll| ll| ll| ll |ll}
		%		\toprule[1pt]
		\hline \hline
		& \multicolumn{2}{c|}{Bicubic} & \multicolumn{2}{c|}{JBF\cite{kopf2007joint}} & \multicolumn{2}{c|}{GF\cite{he2013guided}} & \multicolumn{2}{c|}{SDF\cite{ham2017robust}} & 
		\multicolumn{2}{c|}{DJF\cite{li2016deep}} & 
		\multicolumn{2}{c|}{JFSM\cite{shen2015multispectral}} & \multicolumn{2}{c}{Proposed} \\
		& SSIM & PSNR & SSIM & PSNR & SSIM & PSNR & SSIM & PSNR & SSIM & PSNR & SSIM & PSNR & SSIM & PSNR \\
		\hline
		urban\_0004 & 0.9029 & 25.93 & 0.9359 & 28.47 & 0.9391 & 28.75 & 0.9066 & 26.82 & 0.9789 & 31.02 & 0.9721 & 30.86 & \textbf{0.9811} & \textbf{34.14} \\
		urban\_0006 & 0.9458 & 30.89 & 0.9311 & 32.10 & 0.9400 & 32.66 & 0.8918 & 30.60 & \textbf{0.9894} & 36.04 & 0.9741 & 32.86 & 0.9868 & \textbf{36.79} \\
		urban\_0017 & 0.9527 & 30.45 & 0.9172 & 31.11 & 0.9205 & 31.32 & 0.9281 & 30.72 & \textbf{0.9815} & 34.18 & 0.9500 & 32.85 & 0.9777 & \textbf{35.27} \\
		urban\_0018 & 0.9298 & 25.19 & 0.9308 & 27.59 & 0.9251 & 27.70 & 0.9196 & 26.09 & \textbf{0.9888} & 30.72 & 0.9774 & 30.80 & 0.9874 & \textbf{33.01} \\
		urban\_0020 & 0.9577 & 28.03 & 0.9523 & 30.67 & 0.9494 & 30.69 & 0.9505 & 29.09 & \textbf{0.9915} & 33.60 & 0.9797 & 32.61 & 0.9893 & \textbf{36.66} \\
		urban\_0026 & 0.8704 & 26.27 & 0.8627 & 26.82 & 0.8571 & 26.89 & 0.8558 & 26.61 & 0.9397 & 29.21 & 0.9332 & 28.97 & \textbf{0.9482} & \textbf{30.35} \\
		urban\_0030 & 0.8401 & 26.54 & 0.8476 & 27.58 & 0.8383 & 27.59 & 0.8415 & 27.21 & 0.9345 & 31.27 & 0.9064 & 30.56 & \textbf{0.9443} & \textbf{32.71} \\
		urban\_0050 & 0.9434 & 26.65 & 0.9099 & 27.32 & 0.9116 & 27.35 & 0.9207 & 27.07 & 0.9616 & 28.58 & 0.9251 & 27.58 & \textbf{0.9663} & \textbf{29.37} \\
		\hline
		average & 0.9179 & 27.49 & 0.9109 & 28.96 & 0.9101 & 29.12 & 0.9018 & 28.03 & 0.9707 & 31.83 & 0.9522 & 30.89 & \textbf{0.9726} & \textbf{33.54} \\
		\hline \hline
	\end{tabular}
	\label{Tab:NIR4x_Joint}
\end{table*}

\section{Experiments}
\label{sec:Experiments}

\vspace{-0.2cm}

We now present a series of experiments to validate the effectiveness of the proposed multimodal image processing approaches. The dataset is from the EPFL infrared/RGB image database\footnote{\scriptsize \url{http://ivrl.epfl.ch/supplementary_material/cvpr11/}}. Each pair of near-infrared/RGB images in the dataset has been registered with each other. The target modality is near-infrared and the guidance modality is RGB. 
%As the response of near infrared band exhibit poor correlation with the response of the visible band, it is usually difficult to infer the brightness of an infrared image given a corresponding RGB version. Thus, it is challenging to take good advantage of the RGB modality to aid the processing of the infrared version.
%
%All the images in our dataset are houses and buildings that contain many fine textures and sharp edges. This makes the denoising task more challenging than purifying images with smoother textures, as it is very likely to smooth the sharp details during the denoising process.
%
%
The image pairs are randomly separated into two disjointed groups: training group and testing group. The training image pairs lead to a training dataset consisting of 15000 image patch pairs of $8 \times 8$ pixels. For denoising experiments, we add zero-mean white Gaussian noise with different standard deviations $\sigma = [4,8,12,16,20,24]$ into the near-infrared images to generate the noisy versions. For super-resolution experiments, we blur and downsample each HR near-infrared image by a factor, e.g., 4 $\times$, using the MATLAB "imresize" function to generate corresponding LR versions, similar to \cite{yang2008image,yang2010image}.
%The practical multispectral/RGB datasets are obtained from the Columbia multispectral database\footnote{\url{http://www.cs.columbia.edu/CAVE/databases/multispectral/}}.
%
We compare our approach with state-of-the-art joint image filtering approaches such as~\cite{kopf2007joint,he2013guided,ham2017robust,shen2015multispectral} and deep learning based approach~\cite{li2016deep}. The same guidance images as in our approach are leveraged by these comparison approaches as well.
%
%We compare our approach with state-of-the-art joint image filtering approaches, including Joint Bilateral Filtering (JBF)\cite{kopf2007joint}, Guided image Filtering (GF)\cite{he2013guided}, Static/Dynamic Filtering (SDF)\cite{ham2017robust}, Deep Joint image Filtering (DJF)\cite{li2016deep} and Joint Filtering via optimizing a Scale Map (JFSM)\cite{shen2015multispectral}. The same guidance images as in our approach are leveraged for these comparison approaches.
%and their parameters are tuned to be optimum for the modalities and noise levels.\footnote{The parameters of these approaches have been tuned to produce the best results for the infrared/RGB modalities and different noise levels $\sigma = [4,8,12,16,20,24]$. \\	
%	Parameters for JBF: window size = [7,9,11,13,15,17], $\sigma_d$=[4,5,6,7,8,9], $\sigma_r$ = 0.01.\\
%	Parameters for GF: nhoodSize = [3,5,5,7,9,11], smoothValue  = 1e-5. \\
%	Parameters for SDF: $\lambda$ = 10, $\mu$ = 300, $\nu$ = 200, step=1.\\
%	Parameters for DJF: the net structure keeps intact as in \cite{li2016deep}, i.e., for CNN$_T$ and CNN$_G$: (9x9x3x96) -> (1x1x96x48) -> (5x5x48), for CNN$_F$: (9x9x2x64) -> (1x1x64x32) -> (5x5x32). TrainSize = 50,000 or 160,000; Epochs = 200, batchSize = 256, learningRate = 0.001~0.01, weightDecay = 0.05.\\
%	Parameters for JFSM: $\lambda$ = [14,13,11,9,7,6], $\beta$ = 0.5, maxIter = 5.
%}
We adopt the Peak Signal to Noise Ratio (PSNR), the Root Mean square error (RMSE), and the Structure SIMilarity (SSIM) index\cite{wang2004image} as the image quality evaluation metrics which are commonly used in the image processing literature.

% Residual for Reconstructed Infrared images. comparison with joint SR approaches including\cite{ham2017robust,li2016deep,shen2015multispectral,zhang2014rolling,he2013guided,kopf2007joint}.
\begin{figure}[t]
	\centering
	%---------------------------------------------------
	% im No. 1
	\begin{minipage}[b]{0.24\linewidth}
		\centering
		{\footnotesize urban\_0004} \hfill % \vfill
		\includegraphics[width = 2cm, height= 2cm]{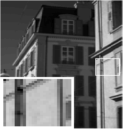}
	\end{minipage} 
	\begin{minipage}[b]{0.24\linewidth}
		\centering
		\includegraphics[width = 2cm, height= 2cm]{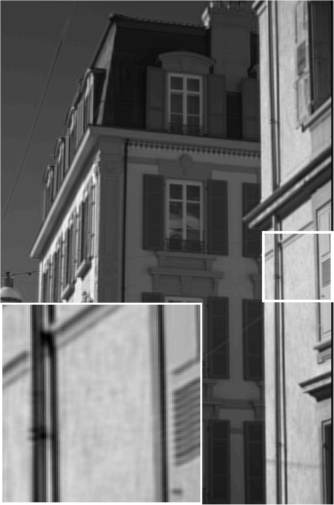}
	\end{minipage} 
	\begin{minipage}[b]{0.24\linewidth}
		\centering
		\includegraphics[width = 2cm, height= 2cm]{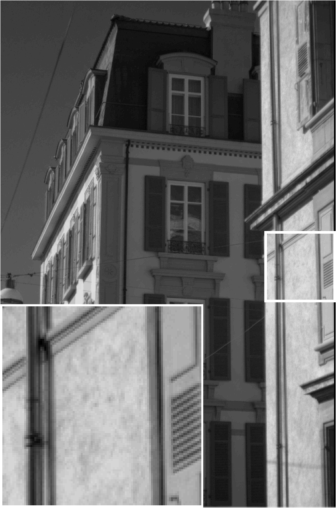}
	\end{minipage} 
	\begin{minipage}[b]{0.24\linewidth}
		\centering
		\includegraphics[width = 2cm, height= 2cm]{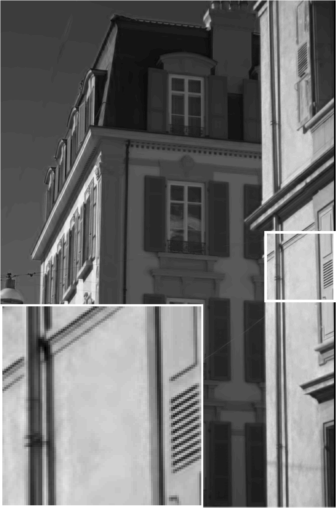}
	\end{minipage} 
	\\
	% -------------------------------------------
	% SI and residue
	%	\begin{minipage}[b]{0.24\linewidth}
	%		\centering
	%		\includegraphics[width = 2cm, height= 2cm]{ImgNo1_SI_true.png}
	%	\end{minipage} 
	\begin{minipage}[b]{0.24\linewidth}
		\centering
		\includegraphics[width = 2cm, height= 2cm]{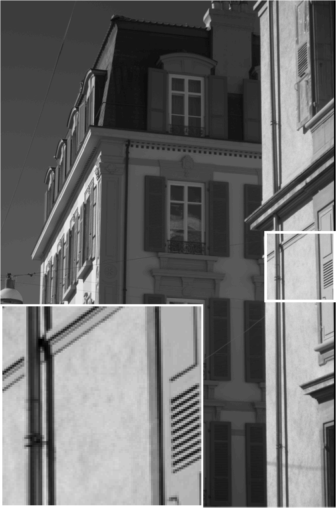}
	\end{minipage} 
	\begin{minipage}[b]{0.24\linewidth}
		\centering
		\includegraphics[width = 2cm, height= 2cm]{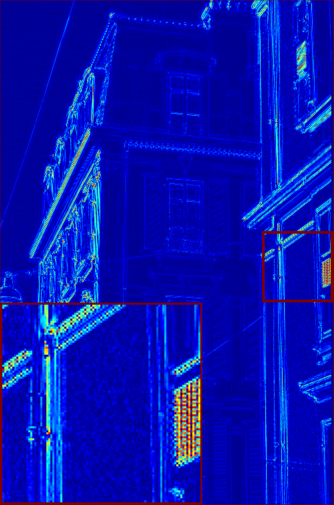}
	\end{minipage} 
	\begin{minipage}[b]{0.24\linewidth}
		\centering
		\includegraphics[width = 2cm, height= 2cm]{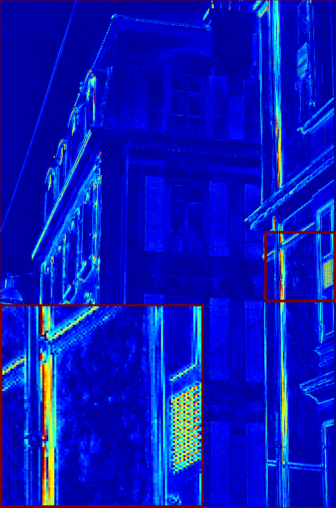}
	\end{minipage} 
	\begin{minipage}[b]{0.24\linewidth}
		\centering
		\includegraphics[width = 2cm, height= 2cm]{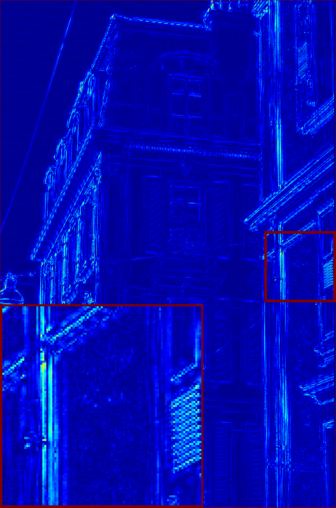}
	\end{minipage} 
	\\
	%---------------------------------------------------
	% im No. 7
	\begin{minipage}[b]{0.24\linewidth}
		\centering
		{\footnotesize urban\_0030} \hfill % \vfill
		\includegraphics[width = 2cm, height= 2cm]{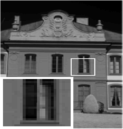}
	\end{minipage} 
	\begin{minipage}[b]{0.24\linewidth}
		\centering
		\includegraphics[width = 2cm, height= 2cm]{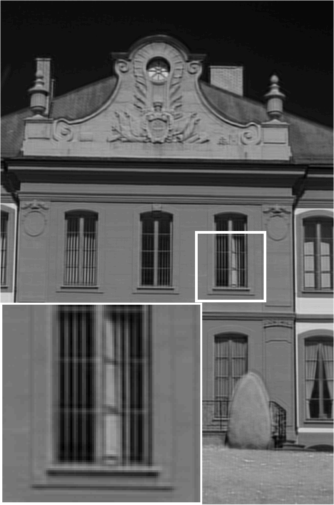}
	\end{minipage} 
	\begin{minipage}[b]{0.24\linewidth}
		\centering
		\includegraphics[width = 2cm, height= 2cm]{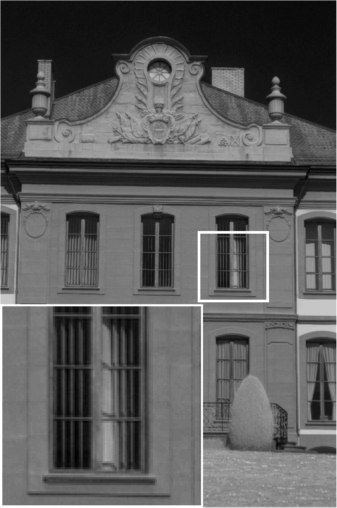}
	\end{minipage} 
	\begin{minipage}[b]{0.24\linewidth}
		\centering
		\includegraphics[width = 2cm, height= 2cm]{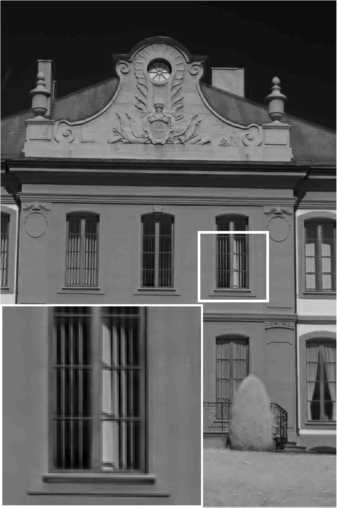}
	\end{minipage} 
	\\
	% -------------------------------------------
	% SI and residue
	%	\begin{minipage}[b]{0.24\linewidth}
	%		\centering
	%		\includegraphics[width = 2cm, height= 2cm]{ImgNo7_SI_true.png}
	%	\end{minipage} 
	\begin{minipage}[b]{0.24\linewidth}
		\centering
		\includegraphics[width = 2cm, height= 2cm]{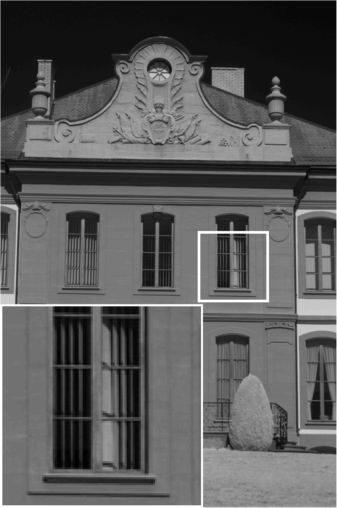}
	\end{minipage} 
	\begin{minipage}[b]{0.24\linewidth}
		\centering
		\includegraphics[width = 2cm, height= 2cm]{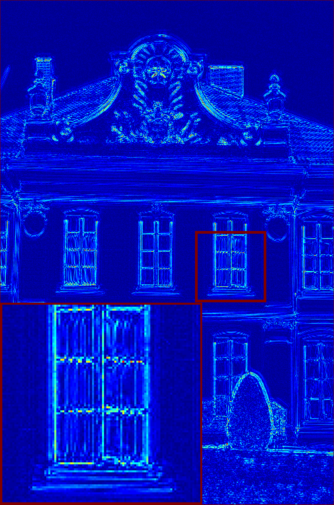}
	\end{minipage} 
	\begin{minipage}[b]{0.24\linewidth}
		\centering
		\includegraphics[width = 2cm, height= 2cm]{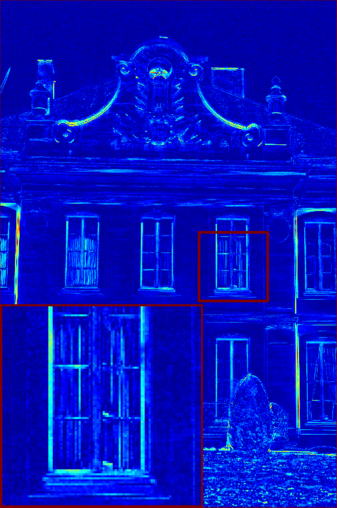}
	\end{minipage} 
	\begin{minipage}[b]{0.24\linewidth}
		\centering
		\includegraphics[width = 2cm, height= 2cm]{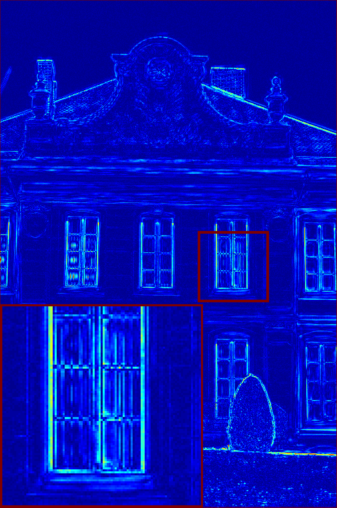}
	\end{minipage} 
	\\
	%---------------------------------------------------
	%	\begin{minipage}[b]{0.08\linewidth}
	%		\centering	{\footnotesize (a)LR}
	%	\end{minipage} 
	\begin{minipage}[b]{0.24\linewidth}
		\centering	{\footnotesize (a) Input/Truth} 
	\end{minipage}
	\begin{minipage}[b]{0.24\linewidth}
		\centering	{\footnotesize (b) DJF\cite{li2016deep}} 
	\end{minipage} 
	\begin{minipage}[b]{0.24\linewidth}
		\centering	{\footnotesize (c) JFSM\cite{shen2015multispectral}} 
	\end{minipage} 
	\begin{minipage}[b]{0.24\linewidth}
		\centering 	{\footnotesize (d) Proposed}
	\end{minipage} 
	%----------------------------------------------------------

	\vspace{-0.3cm}
	
	\begin{multicols}{1}  
		\begin{minipage}[b]{0.98\linewidth}
			%			{\footnotesize Colorbar}
			\includegraphics[width = 8.5cm, height=0.3cm]{Colorbar64-eps-converted-to.pdf}
		\end{minipage}
	\end{multicols}	
	
	\vspace{-0.8cm}
	
	%----------------------------------------------------------
	\caption{Multimodal image super-resolution for near-infrared images (4$\times$ upscaling). For each scene, the first row is the LR input and SR results and the second row is the ground truth and corresponding error map for each approach.}
%		In the error map, brighter area represents larger error.}
	\label{Fig:NIR4x_Joint}
\end{figure}

%% Noise 12 in both training and testing with mismatch
\begin{figure}[t]
	\centering
	% -------------------------------------------
	\begin{minipage}[b]{0.48\linewidth}
		%		\raggedright
		\centering
		\includegraphics[width = 4.2cm]{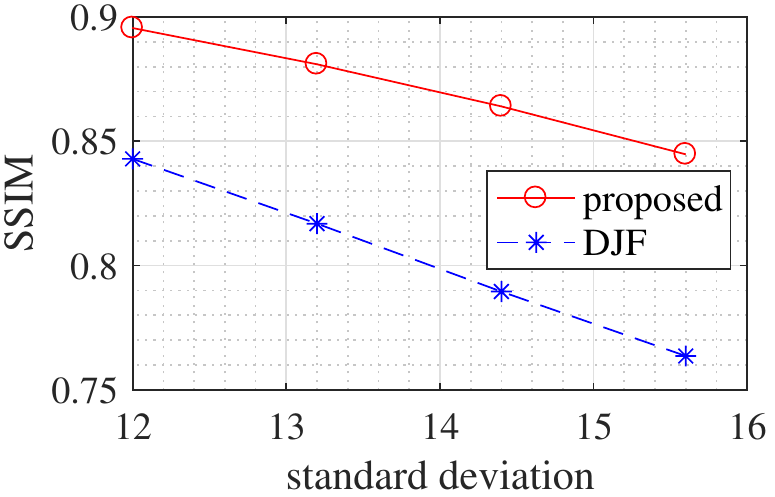} \\
		\scriptsize (a) SSIM w.r.t. Testing Noise
	\end{minipage} 
	% -------------------------------------------
	\begin{minipage}[b]{0.48\linewidth}
		%		\raggedright
		\centering
		\includegraphics[width = 4.2cm]{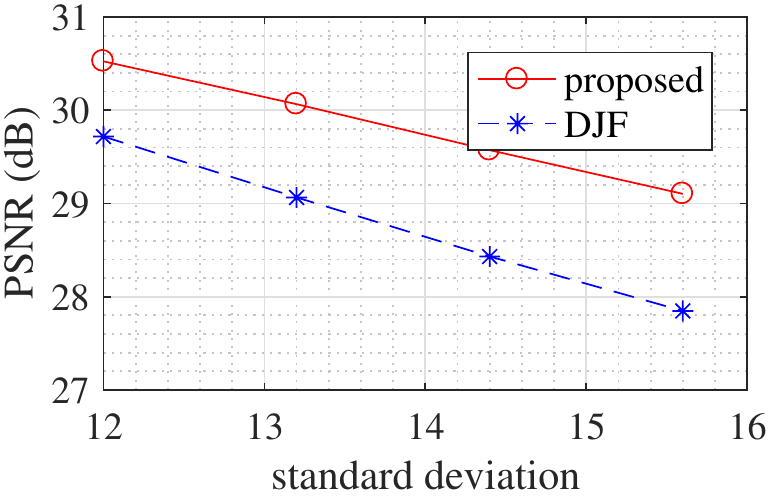} \\
		\scriptsize (b) PSNR w.r.t. Testing Noise
	\end{minipage}
	% -------------------------------------------
	\caption{Both the LR testing and training images are noisy for multimodal image SR (4$\times$ upscaling), using DJF~\cite{li2016deep} and proposed approach. The standard deviation of the noise in the testing images ranges from 12 to 15.6.}
	\label{Fig:NIR4x_Joint_NoiseMismatch_Line}
\end{figure}

%Figure~\ref{Fig:LearnedD} shows the learned coupled dictionaries from the corpus of clean infrared images and corresponding RGB version. We can find that any pair of atoms from common dictionaries $\boldsymbol{\Psi}_{c}$ and $\boldsymbol{\Phi}_{c}$ capture associated edges, blobs, textures with the same direction and location, as well as exhibit considerable resemblance and strong correlation to each other. This outcome indicates that the common dictionaries have indeed captured the similarities between infrared and RGB modalities. In contrast, the learned unique dictionaries $\boldsymbol{\Psi}$ and $\boldsymbol{\Phi}$ represent the disparities of these modalities and therefore rarely exhibit resemblance.

Figure~\ref{Fig:DenoisedIms_7} demonstrates the visual quality of the purified infrared images, as well as the corresponding error maps. As shown in the figure, our approach substantially attenuates the noise, reliably reserves image sharp details and suppresses artifacts. Therefore, the purified near-infrared images by our approach are cleaner and more visually plausible than the reconstruction by the competing approaches. The visual quality is also demonstrated by the error maps where the denoised near-infrared images using our approach exhibit the least residual for different noise levels in comparison with the competing methods. 
%\textcolor{blue}{
	In particular, it can be observed that our approach also outperforms deep learning based approach~DJF\cite{li2016deep}. We believe that the good robustness and stability is due to sparsity priors exploited by our model.
%} 
The average PSNR and average RMSE results for the multimodal image denoising task, shown in Figure~\ref{Fig:PSNR_RMSE}, also confirm that our approach exhibits notable advantages over the competing methods.

Figure~\ref{Fig:NIR4x_Joint} compares the visual quality of the reconstructed HR near-infrared images and the corresponding error maps. It can be seen that our approach recovers more visually plausible images, exhibiting less error than the competing methods, Therefore, our reconstruction is more photo-realistic and visually appealing than the counterparts. Table~\ref{Tab:NIR4x_Joint} also confirm the significant advantage of the proposed approach over other state-of-the-art methods. 
%\textcolor{blue}{
Figure~\ref{Fig:NIR4x_Joint_NoiseMismatch_Line} shows the quantitative SR results for noisy LR input images of target modality. It is noticed that our algorithm demonstrates reasonable stability and robustness to noise, especially to strong noise. In contrast, DJF\cite{li2016deep} is susceptible to noise and its performance degrades faster than ours. 
%}
%Overall, this illustrates that our method is more robust to mismatched noise. 
%We believe that the good robustness and stability is due to sparsity priors exploited by our model.

%In particular, this indicates that detailed structure information can be effectively captured by coupled dictionary learning, especially on images such as buildings and houses that contain a lot of sharp edges, textures and stripes.\footnote{Limited to space, only a few algorithms producing the best results are shown in the paper.} 
%Our approach also outperforms the deep-learning-based approach DJF\cite{li2016deep} for the selected number of training samples.

\vspace{-0.2cm}

\section{Conclusion}
\label{sec:Conclusion}

\vspace{-0.2cm}

%\textcolor{blue}{
This paper proposes a multimodal image processing framework based on coupled dictionary learning. The proposed scheme explicitly incorporates sparsity prior and cross-similarity prior in the data model to captures the similarities and disparities between different image modalities in a learned sparse feature domain in lieu of the original image domain. In this way, the proposed scheme can take better advantage of a guidance image modality to aid the processing of another different image modality in various tasks such as denoising, super-resolution. The experiments demonstrate that our framework achieves notable benefits with respect to the state-of-the-art, as well as outperforms deep-learning-based methods especially when the data is contaminated by noise. 
%This illustrates that our approach has better robustness, but consume much less computing resource and training time.
%}

\clearpage

% % use section* for acknowledowgment
% \section*{Acknowledgment}
% 
% 
% The authors would like to thank...

%% Can use something like this to put references on a page
%% by themselves when using endfloat and the captionsoff option.
%\ifCLASSOPTIONcaptionsoff
%  \newpage
%\fi

% trigger a \newpage just before the given reference
% number - used to balance the columns on the last page
% adjust value as needed - may need to be readjusted if
% the document is modified later
%\IEEEtriggeratref{8}
% The "triggered" command can be changed if desired:
%\IEEEtriggercmd{\enlargethispage{-5in}}

\pagebreak

\end{document}